%% file: DAGGER_main.tex
\documentclass[letterpaper]{article} 
\usepackage{aaai2027}  
\usepackage[hyphens]{url}  
\usepackage{graphicx} 
\usepackage{natbib}  
\usepackage{caption} 
\usepackage{algorithm}
\usepackage{algorithmic}

\usepackage{newfloat}
\usepackage{listings}
\DeclareCaptionStyle{ruled}{labelfont=normalfont,labelsep=colon,strut=off} 
\floatstyle{ruled}
\newfloat{listing}{tb}{lst}{}
\floatname{listing}{Listing}

\usepackage{booktabs}

\usepackage{amsmath,amsthm,amssymb}
\usepackage{comment}
\usepackage{soul}
\usepackage{dblfloatfix}
\usepackage{cuted}
\usepackage{array}
\usepackage{tabularx}

\newcolumntype{Y}{>{\centering\arraybackslash}X}

\usepackage{xcolor}

\definecolor{darkgreen}{RGB}{0,128,0}

\newcommand{\system}{\texttt{DAGGER}}

\newcommand{\edge}{edge}
\newcommand{\sequential}{sequential}
\newcommand{\Sequential}{Sequential}
\newcommand{\fanout}{branching}
\newcommand{\branching}{branching}
\newcommand{\Fanout}{Branching}
\newcommand{\Branching}{Branching}
\newcommand{\generalized}{arbitrary}

\title{{\system}: Deployment-Free Quality Estimation for Generative AI Workflows}
\author {
    Dimitrios Liakopoulos,
    Neeraja J. Yadwadkar
}
\affiliations {
    The University of Texas at Austin\\
    dimliak@utexas.edu, neeraja@austin.utexas.edu
}

\begin{document}

\maketitle

\begin{abstract}
Estimating the quality of generative AI workflows is challenging because these workflows comprise multiple stages, and errors can propagate across them, making end-to-end performance depend not only on the accuracy of individual models but also on how their outputs affect downstream stages.
Each stage can use one of many candidate models. Estimating end-to-end workflow quality by deploying and evaluating every resulting model configuration is therefore prohibitively expensive.
As a result, existing approaches face a tradeoff between estimation cost and accuracy: inexpensive methods often produce unreliable quality estimates, while accurate methods require deploying and evaluating many workflow configurations.
In this paper, we introduce {\system}, a framework for estimating the quality of generative AI workflows without exhaustive deployment and profiling. {\system} represents a workflow as a graph, in which nodes correspond to stages and edges capture how outputs from one stage affect subsequent stages. It then decomposes workflow quality into per-node accuracy terms and per-edge dependency terms. Using a \emph{novel analytical formulation and SVD-based matrix completion}, {\system} estimates workflow quality from sparse observations without executing every candidate workflow.
We evaluate {\system} on three generative AI workflows across three datasets. {\system} estimates workflow quality with a median error of 6.8 percentage points while reducing profiling cost by $10.2\times$ relative to BenchPress, a state-of-the-art baseline, and selects workflow configurations competitive with state-of-the-art frameworks such as LLMSelector. 

\end{abstract}

\input{01_Introduction}
\input{02_Background_Motivation}
\input{03_Formulations}

\input{04_Evaluation}
\input{05_Related}

\input{06_Conclusion}

\bibliography{refs}

\clearpage
\appendix
\input{07_Appendix_A}


\end{document}

%% file: 01_Introduction.tex
\vspace{-10mm}
\section{Introduction}
\label{sec:intro}
\vspace{-3mm}

\begin{figure}[t]
\centering
\includegraphics[width=.75\linewidth]{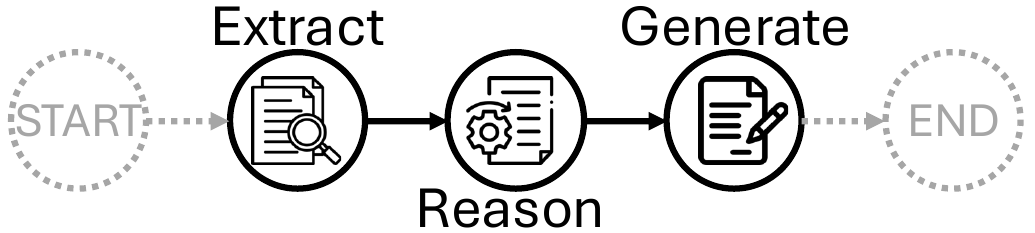}
\vspace{-3mm}
\caption{The graph of a multi-hop question-answering task showing subtasks and their interdependencies. See \S~\ref{sec:intro}.}
\label{fig:intro_example}
\vspace{-5mm}
\end{figure}

Generative AI applications are increasingly built as multi-step workflows where multiple Large Language Models (LLMs) collaborate to accomplish complex tasks such as multi-hop question-answering, retrieval-augmented generation, and agentic reasoning~\cite{BAIR_CompoundAI, AWS_Quick_Flows, Google_Vertex_Agent_Engine, Azure_AI_foundry, LangChain, ScaffoldedLLMs, Murakkab, Nalar, AFlow}. 
These applications are commonly structured as directed acyclic graphs (DAGs), where nodes represent LLM-powered subtasks (or stages) and edges capture information flow between stages.
For example, a multi-hop question-answering pipeline (Figure~\ref{fig:intro_example}) may consist of a retrieval model that extracts relevant passages, a reasoning model that connects evidence across passages, and a generation model that synthesizes the final response.
In such pipelines, imperfect outputs from one stage become the inputs of the next. Thus,
\emph{the quality of such workflows depends not only on the accuracy of individual models, but also on how models interact across stages}~\cite{LLMSelector, LLMChemistry, BAIR_CompoundAI}.

Estimating this end-to-end quality is difficult without evaluating the many model combinations that can instantiate a workflow. Each stage may use one of several candidate models with different accuracy~\cite{FLORA, AgenticPredictor}, latency, and cost~\cite{LLMSelector}. The resulting design space grows combinatorially with the number of stages and candidates. For example, Hugging Face~\cite{HuggingFace} provides roughly 620,000 NLP models with diverse performance characteristics, and evaluating just 27 model combinations for the workflow in Figure~\ref{fig:intro_example} produces a $1.43\times$ variation in quality, from 46.5\% to 66.5\%. So, even modest workflows can yield thousands to millions of possible configurations, making exhaustive execution prohibitively expensive.

\begin{figure}[t]
\centering
\includegraphics[width=.65\linewidth]{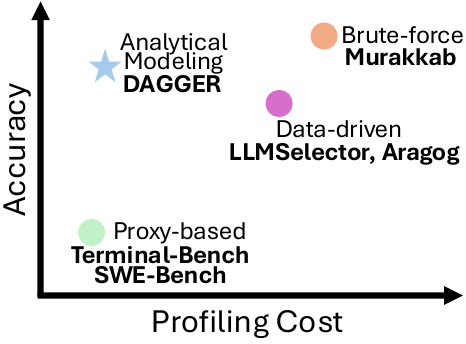}
\vspace{-4mm}
\caption{Current quality estimation approaches present a tradeoff between estimation accuracy and profiling cost. 
Ideally, we require methods with high estimation accuracy and similar cost to proxy-based methods. See \S~\ref{sec:intro} and~\S~\ref{sec:related}.
} 
\label{fig:intro_landscape}
\vspace{-5mm}
\end{figure}

Existing approaches \emph{cannot accurately estimate end-to-end workflow quality without incurring substantial profiling cost}. They primarily follow three strategies. First, benchmark-based methods~\cite{terminalbench, swebench} use proxy metrics derived from standardized evaluations (e.g., Open LLM Leaderboard~\cite{openllmleaderboard}). They are inexpensive, as they require no deployment and reduce quality estimation to a table lookup. But, these proxies can mispredict workflow quality, as the quality depends on the specific model combination and target dataset. Second, exhaustive-profiling methods~\cite{Murakkab} directly evaluate every candidate workflow configuration, yielding accurate estimates at prohibitive cost. 
For example, exhaustively profiling only 27 model combinations in Figure~\ref{fig:intro_example} takes 2.52 hours, requires 52.5 GB of GPU memory, and costs \$16.98 on an RTX PRO 6000 Blackwell cluster.
Third, data-driven pruning methods~\cite{Aragog, LLMSelector} reduce the number of configurations evaluated but still require deploying and measuring a substantial subset of candidate workflows. 
So, existing approaches face a tradeoff between profiling cost and quality-estimation fidelity: inexpensive proxies are often inaccurate, while more accurate methods require substantial deployment and measurement overhead (Figure~\ref{fig:intro_landscape}). \emph{No existing system accurately estimates the end-to-end quality of generative AI workflows from sparse measurements without exhaustively deploying and profiling candidate configurations.}

We address this challenge by exploiting the structure of workflow quality. Prior work shows that end-to-end quality depends on both individual-model accuracy and cross-stage interactions~\cite{LLMSelector, LLMChemistry, BAIR_CompoundAI}. Our key insight is that workflow quality can be decomposed into node-level accuracy and edge-level dependency terms that can be analytically recombined (see~\S~\ref{analytical-model}). We derive a novel formulation (see~\S~\ref{sec:formulation}) that uses these terms to estimate a candidate workflow’s quality without deploying and executing it end to end.

The remaining challenge is obtaining these node- and edge-level measurements without recreating the cost of exhaustive profiling. Measuring every model on every dataset, as well as every pair of models across connected stages, would itself be prohibitively expensive. Instead, we formulate both estimation problems as matrix-completion tasks (see~\S~\ref{subsubsec:motiv_SVD}). {\system} profiles only a sparse subset of model--dataset and inter-model combinations, then uses SVD-based matrix completion~\cite{Papailiopoulos} to infer the missing node-accuracy and edge-dependency values. This approach preserves the information needed for accurate end-to-end estimation while substantially reducing profiling overhead.

In this paper, we introduce {\system}, a framework for estimating the quality of generative AI workflows without exhaustive deployment and profiling. Given a workflow DAG, {\system} uses two types of information: per-node accuracies of candidate models and per-edge dependency scores between models assigned to adjacent stages.
For the workflow in Figure~\ref{fig:intro_example}, these capture the performance of candidate retrieval, reasoning, and generation models, together with retrieval-to-reasoning and reasoning-to-generation dependencies.
These terms do not need to be measured on the target task: {\system} organizes them into structured matrices, observes only a sparse subset of entries (e.g., from prior deployments or existing benchmarks), and recovers the rest via SVD-based matrix completion.
At the core of {\system} is our novel flow-based formulation that composes the node- and edge-level terms over the DAG, enabling {\system} to estimate configurations on the fly without exhaustively executing them. 
We make the following contributions:

\begin{list}{\raisebox{0.25ex}{\scriptsize$\bullet$}}{
    \setlength{\leftmargin}{0.8em}
    \setlength{\labelwidth}{0.5em}
    \setlength{\labelsep}{0.3em}
    \setlength{\itemindent}{0pt}
    \setlength{\itemsep}{0pt}
    \setlength{\topsep}{0pt}
    \setlength{\parsep}{0pt}
    }

\item We formalize the problem of estimating end-to-end quality for generative AI workflows and show that workflow quality can be decomposed into per-node accuracy and inter-node dependency terms.

\item We develop {\system}, a flow-based analytical framework that estimates workflow quality without exhaustively deploying and profiling candidate configurations, and leverages SVD-based matrix completion to infer sparse workflow metadata from a small number of observations.

\item We build and evaluate {\system} on three generative AI workflows and three datasets and show that it estimates end-to-end quality with an error of 6.8 percentage points while lowering profiling cost by 10.2$\times$ compared to state-of-the-art baselines.  

\end{list}

%% file: 02_Background_Motivation.tex
\vspace{-2mm}
\section{Problem Context}
\label{sec:background}
\vspace{-1mm}

\subsection{Workflow Abstractions and Terminology}
\label{subsec:terminology}
\vspace{-1mm}

\noindent \textbf{Task.} A task is a high-level application objective performed by one or more LLMs (e.g., multi-hop question answering).

\begin{figure}[t]
\centering
\includegraphics[width=\linewidth]{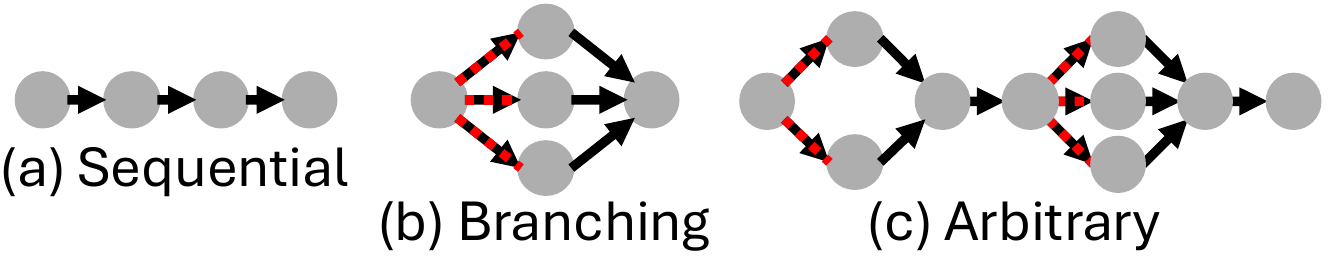}
\vspace{-7mm}
\caption{Representative dependency structures in generative AI workflows: (a) sequential, (b) branching, and (c) arbitrary. Nodes represent models, arrows indicate inter-node dependencies, and red dotted lines illustrate routing across branching paths. See \S~\ref{subsec:terminology}.
}
\label{fig:motive_wfs}
\vspace{-3mm}
\end{figure}

\noindent \textbf{Workflow.} A task is decomposed into multiple subtasks, each corresponding to a specific operation within the application pipeline.
These subtasks form a workflow organized as a DAG, where nodes represent model invocations and edges capture dependencies between stages.
Each subtask can be fulfilled by multiple candidate models, each affecting end-to-end performance differently.
Figure~\ref{fig:motive_wfs} illustrates three representative dependency structures commonly seen in generative AI workflows: (a) {\sequential}, (b) {\branching}, and (c) {\generalized}, which combines sequential and branching patterns.
The red dotted lines in the branching workflow indicate how information is routed across parallel paths (see~\S~\ref{sec:formulation}).

\noindent \textbf{Workflow quality.} 
Following prior work~\cite{mtbench, EvalGen, Prometheus}, we measure the quality of the outputs produced for the downstream task using an LLM-as-a-judge framework, where an LLM evaluator assigns a binary correctness score to the workflow's output.
A workflow's quality is the average score, between (0,1], over a predefined dataset.

\vspace{-2mm}
\subsection{Formal Notation}
\label{subsec:problem_formulation}
\vspace{-1mm}

We represent a generative AI workflow as a directed acyclic graph (DAG), $G = (V, E)$, where each node $v_i \in V$ corresponds to an LLM $M_i$ and each directed edge $(i,j) \in E$ indicates that the output of $M_i$ is consumed by $M_j$. 
In branching workflows, each edge is assigned a weight $w$ that quantifies the relative contribution of the corresponding branch to the overall workflow execution.

\noindent \textbf{Per-Node Accuracy Term.} We define the per-node accuracy term $\alpha_i \in (0,1]$ of model $M_i$ as the probability that $M_i$ produces a correct output when given a correct (oracle) input, measured over dataset $\mathcal{D}$:
\[\alpha_i \;=\; P\!\left(\text{output}_i \text{ is correct} \mid \text{input}_i \text{ is correct};\;\mathcal{D}\right)\]

\noindent \textbf{Per-Edge Dependency Term.} For each edge $(i,j) \in E$, we define the per-edge dependency term $f_{ij} \in (0,+\infty)$ as:
{\fontsize{10.45pt}{12pt}\selectfont\[f_{ij} \;=\; \frac{P\!\left(\text{output}_j \text{ is correct} \mid \text{input}_j = \text{output}_i;\;\mathcal{D}\right)}{\alpha_j}\]}

When $f_{ij} = 1$, model $M_j$ is unaffected by imperfections in the output of $M_i$.
When $f_{ij} < 1$, errors from $M_i$ propagate and degrade the performance of $M_j$.
Finally, when $f_{ij} > 1$, the output of $M_i$ provides richer contextual information than the oracle input alone, improving the effective accuracy of $M_j$.
In all cases, $f_{ij}$ is jointly determined by the model pair $(M_i, M_j)$ and the dataset $\mathcal{D}$ (see~\S~\ref{subsubsec:motiv_SVD}).

\vspace{-2mm}
\section{Characterization and Key Observations}
\label{sec:background_motiv}
\vspace{-1mm}

In this section, we characterize the behavior of generative AI workflows to understand whether end-to-end quality can be estimated analytically without exhaustive deployment and profiling. 
Our measurements lead to two key observations. First, workflow quality exhibits a decomposable structure that enables analytical estimation. Second, sparse observations are sufficient to accurately infer the features required for such estimation. Together, these observations motivate the design of {\system}, enabling us to accurately estimate workflow quality while avoiding the substantial deployment and profiling cost incurred by existing approaches.

\noindent\textbf{Markovian Assumption.}  
We assume that the workflow quality propagation follows the Markov property: for any edge $(i, j)$, the effective accuracy of model $M_j$ depends only on the effective accuracy of its immediate predecessor $M_i$ and the corresponding per-edge dependency term $f_{ij}$, independent of preceding nodes in the workflow DAG. 

Let $\hat{\alpha}_j$ be the effective quality of a workflow ending in node $v_j$. 
For a node $v_j$ with a single incoming edge from $v_i$:
\vspace{-1mm}
$$
\hat{\alpha}_j = \alpha_j \cdot f_{ij} \cdot \hat{\alpha}_i
$$

Our intuition behind this assumption is that
the local dependency structure holds because the per-edge dependency term $f_{ij}$ already captures the interaction between neighboring models $M_i$ and $M_j$. 
Thus, 
the end-to-end quality of a workflow can be estimated using local pairwise quantities without modeling the full workflow jointly (see~\S~\ref{analytical-model}).

\vspace{-2mm}
\subsection{Observation \#1: Workflow Quality is Analytically Decomposable}
\label{analytical-model}
\vspace{-1mm}

Workflow quality depends on both individual-model accuracy and error propagation across stages~\cite{LLMSelector,multiagentllmsystemsfail}.
Existing approaches typically capture these effects by deploying and evaluating each candidate workflow end to end~\cite{Murakkab, LLMSelector}.
We instead ask whether workflow quality can be analytically derived from a small set of measured components. To investigate this question, we conduct a characterization study on a two-GPU cluster with NVIDIA RTX PRO 6000 Blackwell GPUs and three state-of-the-art LLMs: Qwen-2.5-7B, Mistral-7B, and Llama-3.1-8B. We study {\sequential} and {\fanout} workflows for the multi-hop question-answering task in Figure~\ref{fig:intro_example}, using the HotpotQA dataset~\cite{hotpotqa}.

\begin{table}[t]
\small
\centering
\caption{Measured ($\hat{\alpha}_{TRUE}$) and estimated ($\hat{\alpha}_{CALC}$) workflow quality for the two studied workflows. See \S~\ref{analytical-model}.} 
\vspace{-3mm}
\label{tab:motv_results}
\begin{tabular}{lccc}
\toprule
\textbf{Workflow} & $\hat{\boldsymbol{\alpha}}_{\textbf{TRUE}}$ & $\hat{\boldsymbol{\alpha}}_{\textbf{CALC}}$ & \textbf{Abs. Error} \\
\midrule
Sequential & $64.50\%$ & $64.94\%$ & $0.44\%$ \\
Branching    & $64.50\%$ & $69.45\%$ & $4.95\%$ \\
\bottomrule
\end{tabular}
\vspace{-2mm}
\end{table}

\begin{figure}[t]
\centering
\includegraphics[width=.85\linewidth]{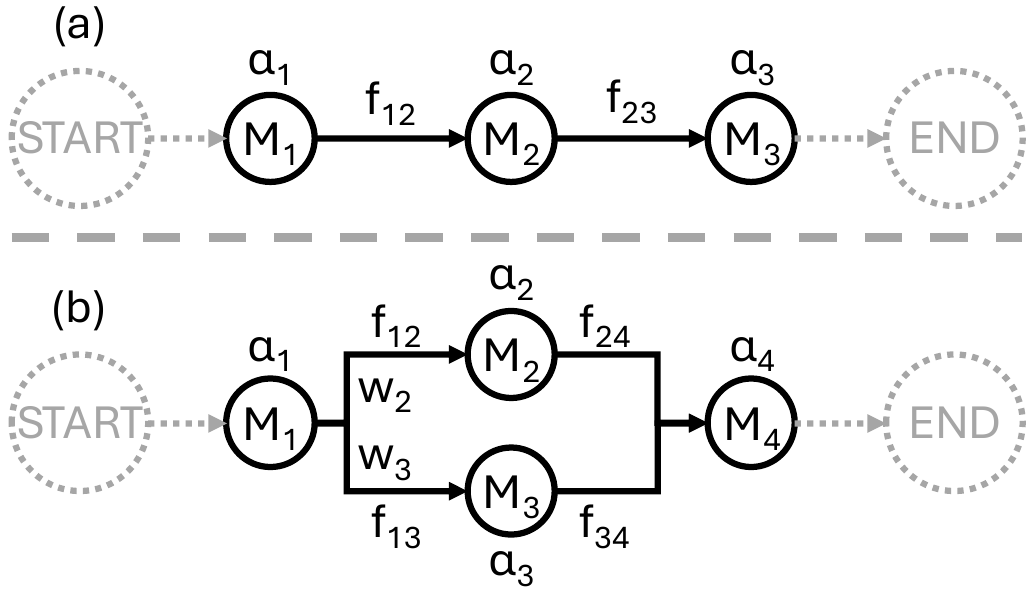}
\vspace{-3mm}
\caption{(a) A sequential and (b) a branching workflow for a multi-hop question-answering task. $M_i$ denotes a model with accuracy $\alpha_i$, while $f_{ij}$ shows the dependency between neighboring models $M_i$ and $M_j$. In the branching workflow, $w_i$ are the weights assigned to each branch. See \S~\ref{sec:background_motiv}, \ref{subsec:formulation_sequential}, \ref{subsec:formulation_fanout}. 
}
\label{fig:motiv_workflows}
\vspace{-6mm}
\end{figure}

\noindent \textbf{{\Sequential} Workflow.} 
We first test whether the quality of a sequential workflow can be decomposed into contributions from its nodes and edges. Consider the three-stage workflow $Extract \rightarrow Reason \rightarrow Generate$ shown in Figure~\ref{fig:motiv_workflows}a, with $M_1$ (Qwen-2.5-7B), $M_2$ (Mistral-7B), and $M_3$ (Llama-3.1-8B). For each node, we measure the accuracy $\alpha_i$ of model $M_i$ on its assigned subtask using an LLM-based evaluator. For each edge, we measure a dependency term $f_{ij}$ that captures how the output quality of $M_i$ affects the downstream performance of $M_j$. We then estimate the end-to-end workflow quality as: 
$\hat\alpha_{CALC} = \alpha_1 \cdot f_{12} \cdot \alpha_2 \cdot f_{23} \cdot \alpha_3$.
This formulation follows the workflow path, combining each model’s individual accuracy with the dependency between consecutive stages. The resulting estimate closely matches the measured end-to-end quality, $\hat{\alpha}_{TRUE}$, with only $0.44\%$ error (Table~\ref{tab:motv_results}).

\noindent \textbf{{\Branching} Workflow.}
Next, we investigate whether the same observation holds for workflows with branching structures.
We extend the previous {\sequential} workflow by introducing two parallel reasoning branches, resulting in the {\branching} workflow shown in Figure~\ref{fig:motiv_workflows}b.
The workflow DAG consists of four nodes: $M_1$ (Qwen-2.5-7B), $M_2$ and $M_3$ (Mistral-7B), and $M_4$ (Llama-3.1-8B).
To simplify the analysis, we use identical models for the two reasoning branches and assign equal branch weights ($w_2 = w_3 = 0.5$).
We observe that the true workflow quality, $\hat{\alpha}_{TRUE}$, can be closely approximated by:
$\hat\alpha_{CALC} = \alpha_1 \cdot \frac{w_2 \cdot \alpha_2 \cdot f_{12} \cdot f_{24} + w_3 \cdot \alpha_3 \cdot f_{13} \cdot f_{34}}{w_2 + w_3} \cdot \alpha_4$.
This formulation extends the sequential decomposition by incorporating branch weights across parallel paths.
Our measurements show that the estimated quality tracks the measured quality closely, with an error of $4.95\%$ (Table~\ref{tab:motv_results}).

\vspace{1mm}
\noindent
\begingroup
  \setlength{\fboxsep}{4pt}
  \setlength{\fboxrule}{0.05mm}
  \hspace{-2.5mm}
  \makebox[\linewidth][l]{%
    \fcolorbox{black}{gray!10}{%
      \parbox{\dimexpr\linewidth-2\fboxsep-2\fboxrule\relax}{%
        Overall, these results suggest that the end-to-end quality of generative AI workflows can be approximated analytically using per-node accuracy terms, per-edge dependency terms, and branch weights for branching workflows.
      }%
    }%
  }%
\endgroup

\vspace{-2mm}
\subsection{Observation \#2: Sparse Observations are Sufficient for Quality Estimation}
\label{subsubsec:motiv_SVD}
\vspace{-1mm}

We showed that the end-to-end quality of generative AI workflows can be approximated analytically using per-node accuracy terms and per-edge dependency terms.
However, obtaining these terms still requires profiling multiple models across different datasets, resulting in substantial profiling overhead.
Since both per-node and per-edge terms are dataset-dependent, exhaustively collecting them for all model--dataset combinations is prohibitively expensive.
We therefore investigate whether accurate workflow estimation can 
be achieved from only a sparse set of observations.

\noindent \textbf{Per-Node Accuracy Terms.} Prior work~\cite{Papailiopoulos} has shown that LLM benchmarking exhibits strong low-rank structure: only a small fraction of model--benchmark pairs must be evaluated explicitly, while the remaining entries can be inferred through SVD-based matrix completion.
This structure emerges because model families often share architectural similarities and many benchmarks evaluate overlapping capabilities.
We leverage this observation to cast the estimation of per-node accuracy terms, $\alpha_i$, as a matrix completion problem, substantially reducing the profiling cost required to estimate workflow quality.

\noindent \textbf{Per-Edge Dependency Terms.}
The dependency between neighboring models is both model-combination- and dataset-specific.
However, different model combinations often share architectural similarities, while many datasets evaluate overlapping capabilities.
So, we hypothesize that the matrix of per-edge dependency terms also exhibits strong low-rank structure, similarly to prior observations for LLM benchmarking matrices~\cite{Papailiopoulos}.
If this hypothesis holds, then only a sparse subset of dependency values must be profiled explicitly, while the remaining entries can be inferred through SVD-based matrix completion.

To validate this hypothesis, we construct simple two-stage workflows and evaluate them across three datasets (HotpotQA, MuSiQue, and 2WikiMultiHopQA) 
using three models: 
$M_1$ (Qwen-2.5-7B), $M_2$ (Mistral-7B), and $M_3$ (Llama-3.1-8B).
We combine these models into the following model combinations:
$MC_1$:$M_1$$\rightarrow$$M_2$, 
$MC_2$:$M_2$$\rightarrow$$M_1$, 
$MC_3$:$M_1$$\rightarrow$$M_3$, 
$MC_4$:$M_3$$\rightarrow$$M_1$, 
$MC_5$:$M_2$$\rightarrow$$M_3$, 
and $MC_6$:$M_3$$\rightarrow$$M_2$.
For each model combination and dataset pair, we measure the true per-edge dependency term.
We then randomly sample $9$ entries from the resulting dependency matrix and use SVD-based matrix completion to estimate the remaining values.
Table~\ref{tab:motiv_dependency} summarizes the measured and estimated dependency terms, where the sampled entries used for matrix completion are highlighted in red.
Our results show that the estimated dependency values closely match the measured ones, with an average error of $5.9\%$ ($0.2$--$16.3\%$) 
across all model combinations and datasets.

\vspace{1mm}
\noindent
\begingroup
  \setlength{\fboxsep}{4pt}
  \setlength{\fboxrule}{0.05mm}
  \hspace{-2.5mm}
  \makebox[\linewidth][l]{%
    \fcolorbox{black}{gray!10}{%
      \parbox{\dimexpr\linewidth-2\fboxsep-2\fboxrule\relax}{%
        This observation validates our hypothesis that inter-node dependency matrices exhibit strong low-rank structure and can thus be efficiently approximated through SVD-based matrix completion.
      }%
    }%
  }%
\endgroup

%% file: 03_Formulations.tex
\vspace{-1mm}
\section{Analytical Formulations for Estimating Workflow Quality}
\label{sec:formulation}
\vspace{-1mm}

\begin{table}[t]
\small
\centering
\setlength{\tabcolsep}{1.2pt}
\caption{Measured and estimated inter-node dependency values across model combinations and datasets. 
Red entries are the sampled inputs used for the matrix completion.
See \S~\ref{subsubsec:motiv_SVD}.}
\label{tab:motiv_dependency}
\vspace{-3mm}
\begin{tabular}{l|c|c|c}
\toprule
\multicolumn{4}{c}{$f_{TRUE}/f_{CALC} \; (RelativeError\%)$} \\
\midrule
 & \textbf{Dataset 1} & \textbf{Dataset 2} & \textbf{Dataset 3} \\
\midrule
$\boldsymbol{MC_1}$ & \textcolor{red}{$0.91/0.91$ ($-$)} & \textcolor{red}{$0.63/0.63$ ($-$)} & $0.90/0.85$ ($5.7\%$) \\
$\boldsymbol{MC_2}$ & \textcolor{red}{$0.72/0.72$ ($-$)} & $0.39/0.39$ ($0.3\%$) & \textcolor{red}{$0.63/0.63$ ($-$)} \\
$\boldsymbol{MC_3}$ & $0.85/0.92$ ($7.5\%$) & $0.54/0.63$ ($16.3\%$) & \textcolor{red}{$0.85/0.85$ ($-$)} \\
$\boldsymbol{MC_4}$ & $0.89/0.89$ ($0.2\%$) & $0.60/0.60$ ($0.5\%$) & \textcolor{red}{$0.82/0.82$ ($-$)} \\
$\boldsymbol{MC_5}$ & $0.69/0.72$ ($3.7\%$) & \textcolor{red}{$0.39/0.39$ ($-$)} & $0.69/0.63$ ($7.9\%$) \\
$\boldsymbol{MC_6}$ & \textcolor{red}{$0.90/0.90$ ($-$)} & \textcolor{red}{$0.60/0.60$ ($-$)} & $0.92/0.83$ ($10.5\%$) \\
\bottomrule
\end{tabular}
\vspace{-4mm}
\end{table}

Building on the observations from Section~\ref{sec:background_motiv}, {\system} estimates the end-to-end quality of generative AI workflows without exhaustively deploying and executing candidate models.
In this section, we formalize workflow quality estimation using per-node accuracy terms, per-edge dependency terms, and branch weights.
We first derive the formulation for {\sequential} workflows, then extend it to {\branching} structures, and lastly generalize to arbitrary workflow DAGs.

\vspace{-2mm}
\subsection{Formulation for {\Sequential} Workflows}
\label{subsec:formulation_sequential}
\vspace{-1mm}

We first consider a simple three-stage {\sequential} workflow (see~\S~\ref{analytical-model}, Figure~\ref{fig:motiv_workflows}a): $M_1 \rightarrow M_2 \rightarrow M_3$. Using the Markov property, the end-to-end quality can be expressed as:
\vspace{-1mm}
\[Quality = \alpha_1 \cdot f_{12} \cdot \alpha_2 \cdot f_{23} \cdot \alpha_3\]
\vspace{-1mm}
where $\alpha_i$ denotes the accuracy of model $M_i$ and $f_{ij}$ the dependency between models $M_i$ and $M_j$.

\noindent \textbf{Sequential Chain of $n$ Nodes.} We next extend this formulation to an arbitrary {\sequential} chain of $n$ nodes. 
For a linear workflow $M_1 \rightarrow M_2 \rightarrow \dots \rightarrow M_n$, applying the Markov property recursively yields:
\begin{equation}
\boxed{ Quality_{seq} = \prod_{i=1}^{n} \alpha_i \;\cdot\; \prod_{k=1}^{n-1} f_{k,\,k+1}}
\label{eq:seq_accuracy}
\end{equation}
Thus, given the accuracies $\alpha_i$ of all models $M_i$ and the {\edge} functions $f_{k,k+1}$ between neighbors ($M_k$, $M_{k+1}$), we can estimate the end-to-end quality of any {\sequential} workflow.

\vspace{-2mm}
\subsection{Formulation for {\Fanout} Workflows}
\label{subsec:formulation_fanout}
\vspace{-1mm}

We next consider a simple {\branching} workflow (see~\S\ref{analytical-model}, Figure~\ref{fig:motiv_workflows}b), where model $M_1$ distributes inputs across two parallel branches $M_2$ and $M_3$, whose outputs are propagated to $M_4$.
We assign non-negative branch weights $w_2$ and $w_3$ to the outgoing edges from $M_1$, representing the fraction of inputs routed to branches $M_2$ and $M_3$, respectively. 
This workflow contains two directed paths: $[M_1, M_2, M_4]$ and $[M_1, M_3, M_4]$.
Using the sequential formulation for each branch, the end-to-end workflow quality can be 
written as:
\vspace{-1mm}
{\fontsize{10pt}{12pt}\selectfont\[Quality = \alpha_1 \cdot \frac{w_2 \cdot \alpha_2 \cdot f_{12} \cdot f_{24} + w_3 \cdot \alpha_3 \cdot f_{13} \cdot f_{34}}{w_2 + w_3} \cdot \alpha_4\]}
This formulation extends the sequential workflow model by incorporating branch weights across parallel paths.

\noindent \textbf{{\Branching} Edge Weights.}
For a {\branching} node $v_i$ with multiple outgoing edges ($E(i)$), each edge $(i,j)$ is assigned a non-negative branch weight $w_{j} \ge 0$.
The normalized contribution of branch $(i,j)$ is computed as:
\vspace{-1mm}
\begin{equation}
r_{ij} \;=\; \frac{w_{j}}{\displaystyle\sum_{k:\,(i,k)\in E(i)} w_{k}}
\label{eq:routing}
\vspace{-1mm}
\end{equation}

For {\sequential} nodes with a single outgoing edge, $r_{ij}=1$.

\noindent \textbf{Workflows with $n$ Branches.} 
We next extend the formulation to arbitrary {\branching} workflows.
Consider a node $M_s$ connected to $n$ parallel branches $M_{b_1}, \ldots, M_{b_n}$, whose outputs are aggregated by node $M_t$.
Given branch weights $w_{b_1}, \ldots, w_{b_n}$, the end-to-end 
quality can be expressed as:
\vspace{-1mm}
\begin{equation}
\fontsize{9.5pt}{12pt}\selectfont
\boxed{Quality_{branch} = \alpha_s \cdot \alpha_t \cdot \sum_{k=1}^{n} r_{sb_k} \cdot \alpha_{b_k} \cdot f_{sb_k} \cdot f_{b_kt}}
\label{eq:fanout_accuracy}
\end{equation}
where $\alpha_i$ denotes the per-node accuracy term of model $M_i$, $f_{ij}$ denotes the per-edge dependency term between neighboring models $(M_i, M_j)$, and
$r_{sb_k} = w_{b_k} / \sum_{l} w_{b_l}$ denotes the routing fraction of branch $M_{b_k}$.

\vspace{-2mm}
\subsection{Formulation for Arbitrary Workflows}
\label{subsec:formulation_general}
\vspace{-1mm}

\begin{figure}[t]
\centering
\includegraphics[width=\linewidth]{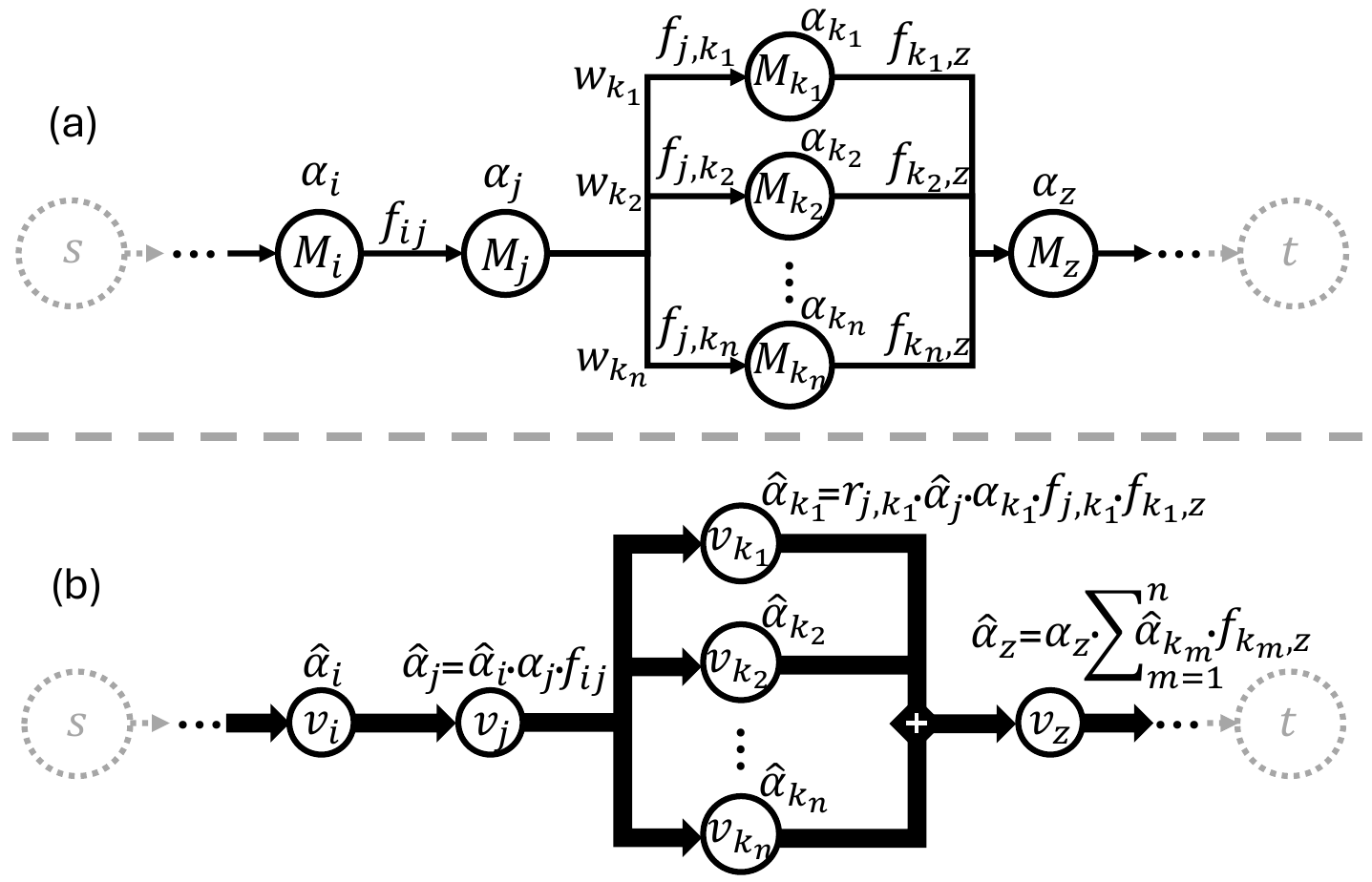}
\vspace{-7mm}
\caption{(a) An {\generalized} workflow consisting of {\sequential} chains and branches for a generative AI task.
$M_i$ denotes a model with per-node accuracy $\alpha_i$, while $f_{ij}$ is the per-edge dependency term between neighboring models $M_i$ and $M_j$. In {\branching} structures, $w_{ij}$ denotes the branch weight assigned to edge $(i,j)$.
(b) The corresponding flow representation of the {\generalized} workflow, where $\hat{\alpha}_i$ denotes the effective workflow quality after traversing node $v_i$ and $r_{ij}$ denotes the normalized contribution of edge $(i,j)$. See \S~\ref{subsec:formulation_general}.}
\label{fig:general_workflow}
\end{figure}

We now generalize the formulation to arbitrary workflow DAGs.
We model a workflow DAG as a flow network in which quality is propagated through the graph and modulated by the contribution of each node and edge (Figure~\ref{fig:general_workflow}).
Each node is associated with a per-node accuracy term $\alpha_i$, while each edge $(i,j)$ has a per-edge dependency term $f_{ij}$.
For {\branching} nodes, the propagated quality is additionally weighted by the normalized branch contribution $r_{ij}$ defined in Equation~\ref{eq:routing}.

\noindent \textbf{Analytical Formulation.}
Let $\mathcal{P}(s,t)$ denote the set of all directed paths from source node $s$ to sink node $t$,
while $\mathcal{V}(p)$ and $\mathcal{E}(p)$ denote the vertices and edges of path $p$. 
For each edge $(i,j)$, we define the edge contribution term:
\vspace{-1mm}
\begin{equation}
\Phi_{ij} =
\begin{cases}
f_{ij} \cdot r_{ij} & \text{for a {\fanout} node } i \\
f_{ij} & \text{otherwise}
\end{cases}
\label{eq:edge_factor}
\vspace{-1mm}
\end{equation}

By combining Equations~\ref{eq:seq_accuracy},~\ref{eq:fanout_accuracy},~\ref{eq:edge_factor}, we can describe the end-to-end quality over the DAG as follows:
\[
\fontsize{10.45pt}{12pt}\selectfont
\boxed{Quality(s,t) = \sum_{p \,\in\, \mathcal{P}(s,t)} \left[\prod_{v \in \mathcal{V}(p)} \alpha_v \right] \left[\prod_{(i,j) \in \mathcal{E}(p)} \Phi_{ij}\right]}
\]
where $\alpha_v$ denotes the per-node accuracy term of model $M_v$ and $\Phi_{ij}$ denotes the edge contribution term defined in Equation~\ref{eq:edge_factor}.
This formulation unifies both {\sequential} and {\branching} workflows, enabling end-to-end quality estimation for arbitrary generative AI workflow DAGs.

%% file: 04_Evaluation.tex
\vspace{-2mm}
\section{Evaluation}
\label{sec:eval}
\vspace{-1mm}

\begin{figure}[t]
\centering
\includegraphics[width=0.8\linewidth]{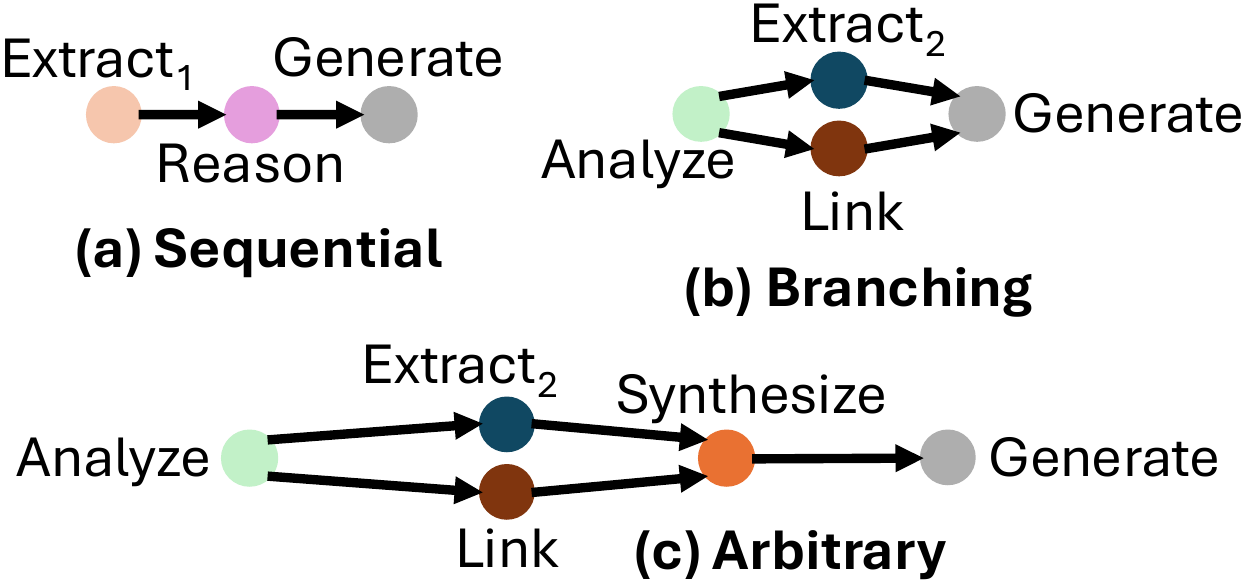}
\vspace{-3mm}
\caption{Workflows used throughout the evaluation. Different color nodes indicate unique operations. See~\S~\ref{sec:eval}.}
\label{fig:eval_wfs}
\vspace{-4mm}
\end{figure}

\subsection{Implementation}
\label{sec:implementation}
\vspace{-1mm}

We build {\system} in Python atop Accelerate~\cite{Accelerate}, a HuggingFace~\cite{HuggingFace} inference serving framework that leverages Fully Sharded Data Parallel (FSDP)~\cite{FSDP} to partition LLMs across GPUs.
For SVD-based matrix completion, we extend BenchPress~\cite{Papailiopoulos} to support both per-node accuracy terms and per-edge dependency terms.
Finally, we use OpenAI's \texttt{gpt-4.1-nano}~\cite{openai_gpt41_nano} as the LLM evaluator, accessed through the OpenAI API.

\subsection{Experimental Setup}
\label{subsec:eval_setup}
\vspace{-1mm}

\noindent \textbf{Workload.} We construct our workload using three generative AI workflows: (a) {\sequential}, (b) {\branching}, and (c) {\generalized}, shown in Figure~\ref{fig:eval_wfs}.
All workflows perform multi-hop question answering by decomposing the task into multiple stages.
We evaluate the workflows using three state-of-the-art LLMs: Qwen-2.5-7B~\cite{qwen-2.5}, Mistral-7B~\cite{mistral}, and Llama-3.1-8B~\cite{llama-3}, across three multi-hop question answering benchmarks: HotpotQA~\cite{hotpotqa}, MuSiQue~\cite{MuSiQue}, and 2WikiMultiHopQA~\cite{2WikiMultiHopQA}.
We evaluate all workflows across all model combinations and datasets.

\noindent \textbf{Baselines.} 
We compare {\system} against four baselines:

(1) \textit{Brute-Force} follows the methodology of prior works (e.g., Murakkab~\cite{Murakkab}), which exhaustively deploy the full workflow under every candidate model combination and capture profiling metrics used during inference.

(2) \textit{Proxy-based} approaches estimate workflow quality using publicly available benchmark scores.
To construct this baseline, we use the average performance across the six benchmarks from Open LLM Leaderboard~\cite{openllmleaderboard} and approximate the workflow quality as the average of the corresponding per-model accuracies.

(3) \textit{BenchPress}~\cite{Papailiopoulos} estimates missing LLM benchmark scores via low-rank SVD-based matrix completion over a [models $\times$ benchmarks] matrix. We adapt it to our setting by treating workflow configurations as rows: for each workflow structure, we construct a [model-combinations $\times$ datasets] matrix of end-to-end quality, observe 33\% of the cells, and reconstruct the remainder using the paper's default predictor.

(4) \textit{LLM-Selector}~\cite{LLMSelector} is a state-of-the-art compound AI framework that uses a data-driven approach for selecting model combinations that maximize end-to-end workflow quality.
We extend LLM-Selector with our evaluation pipeline to find the best workflow configuration for each workflow structure and measure its corresponding quality.

\noindent \textbf{Testbed.} 
We use a server with 2$\times$NVIDIA RTX PRO 6000 Blackwell GPUs and a 32-core AMD EPYC 9354 CPU, running Ubuntu 24.04 with CUDA 13.2 (Driver: 595.58.03).

\noindent \textbf{Metrics.}
We evaluate {\system} on both estimation fidelity and cost efficiency.
To quantify estimation quality, we compare the estimated workflow quality against the measured end-to-end quality obtained using an LLM-as-a-judge evaluator and report the absolute error.
To quantify cost efficiency, we measure profiling overhead in terms of profiling time and GPU-hours following prior work~\cite{MaverIQ}.

\vspace{-2mm}
\subsection{Quality estimation accuracy of {\system}}
\label{subsec:eval_accuracy}
\vspace{-1mm}
{\system} uses SVD-based matrix completion to estimate per-node accuracy and per-edge dependency terms. We first evaluate the error introduced by this step.
We then evaluate the fidelity of {\system}'s analytical formulations in estimating end-to-end workflow quality compared to the baselines.

\noindent \textbf{Per-Node Accuracy Matrix Completion.}
Our workload consists of three models, three datasets, and three workflows that collectively execute seven unique operations (Figure~\ref{fig:eval_wfs}), resulting in 21 dataset-specific operations requiring per-node accuracy profiling.
To reduce this overhead, we randomly sample 53 entries from 63 total and apply SVD-based matrix completion (see~\S~\ref{subsubsec:motiv_SVD}) to estimate the remaining values.
The results for all models and dataset-specific operations are presented in Appendix~\ref{sec:appendix_a} in Table~\ref{tab:eval_pernode_ALL}.
Overall, SVD-based matrix completion achieves a median relative estimation error of 19.1\%, demonstrating that per-node accuracies can be estimated with reasonable fidelity.
We note that the results of the SVD-based matrix completion are greatly affected by the number of observed values per row and column. Including more models (i.e., extending the matrix's rows) would allow us to achieve better or similar estimation with a smaller number of observations~\cite{candes2009exact,JMLR_matrixcompletion}.

\noindent \textbf{Per-Edge Dependency Matrix Completion.}
In addition to per-node accuracy terms, {\system} also requires per-edge dependency terms.
Since workflow configurations may assign arbitrary models to workflow operations, dependency values must be collected across all model combinations and dataset-specific edge connections.
To reduce this overhead, we randomly sample 146 entries from 243 and apply SVD-based matrix completion (see~\S~\ref{subsubsec:motiv_SVD}) to estimate the remaining dependency values.
The results for all model combinations and dataset-specific edge connections are presented in  Table~\ref{tab:eval_peredge_ALL} in Appendix~\ref{sec:appendix_a}.
Overall, SVD-based matrix completion achieves a median relative estimation error of 16.4\%. Thus, the per-edge dependency terms can be estimated accurately from sparse observations.

\begin{table}[t]
\centering
\small
\caption{{\system}'s median estimation error in estimating workflow quality across three workflows and three datasets: $D_1$ (HotpotQA), $D_2$ (MuSiQue), and $D_3$ (2WikiMultiHopQA) for all model combinations. Lower values indicate higher estimation fidelity.
See~\S~\ref{subsec:eval_accuracy}.}
\vspace{-3mm}
\label{tab:eval_accfull}
\begin{tabular}{lccc||c}
\toprule
\textbf{Workflow} & $\boldsymbol{D_1}$ & $\boldsymbol{D_2}$ & $\boldsymbol{D_3}$ & \textbf{Cumulative} \\
\midrule
Sequential & $3.20\%$ & $3.82\%$ & $4.72\%$ & $3.82\%$ \\
Branching  & $8.05\%$ & $2.91\%$ & $5.53\%$ & $5.30\%$ \\
Arbitrary  & $8.59\%$ & $7.70\%$ & $7.58\%$ & $7.84\%$ \\
\midrule
\midrule
Combined & $8.16\%$ & $6.02\%$ & $6.42\%$ & $6.79\%$ \\
\bottomrule
\end{tabular}
\vspace{-3mm}
\end{table}

\noindent \textbf{End-to-End Workflow Quality.}
Using the profiled per-node accuracy terms and per-edge dependency terms, we next evaluate the fidelity of {\system} in estimating end-to-end workflow quality.
For each workflow configuration in Figure~\ref{fig:eval_wfs}, {\system} estimates workflow quality using the analytical formulations described in~\S~\ref{sec:formulation}.
To quantify estimation error, we additionally execute the full workflows and measure the true end-to-end quality as ground truth.
Table~\ref{tab:eval_accfull} shows the error in estimating workflow quality across all workflows, model combinations, and datasets.
Across all settings, {\system} achieves an estimation error of $2.91-8.59\%$ ($6.79\%$ median), demonstrating that the analytical formulations estimate workflow quality with high fidelity. 
We note that {\system}'s fidelity depends on the accuracy and dependency terms recovered by the SVD-based matrix completion, which we initialize at random. Principled initialization is orthogonal to this paper and left to future work.

\begin{figure}[t]
\centering
\includegraphics[width=\linewidth]{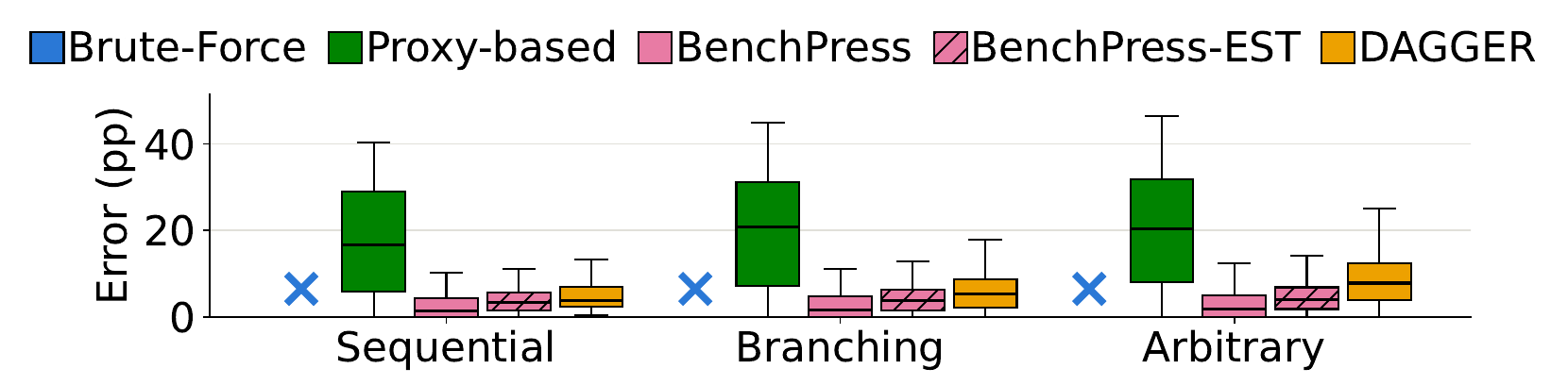}
\vspace{-6mm}
\caption{End-to-end estimation error across model combinations and datasets for the three workflows. {\system} estimates the workflow's quality with minimal errors,
while incurring substantially lower costs (Figure~\ref{fig:eval_profcost}) compared to the baselines, showing that the analytical formulations are highly accurate and cost-effective. 
See~\S~\ref{subsec:eval_accuracy}.}
\label{fig:eval_accError}
\vspace{-4mm}
\end{figure}

\noindent \textbf{{\system} vs Baselines.}
We compare estimation quality against the baselines for the workflow configurations in Figure~\ref{fig:eval_wfs}, excluding LLMSelector as it cannot estimate quality for given workflows.
Figure~\ref{fig:eval_accError} shows the estimation error across all model combinations and datasets per workflow structure.
{\system}'s analytical formulation accurately estimates end-to-end quality, achieving a median error of 6.79 percentage points (pp).
Brute-Force constitutes the ground truth (0 pp error), while BenchPress retains exact ground-truth quality values for 1/3 of the dataset, yielding a 1.72 pp median error; excluding the observed values (BenchPress-EST), the error nearly doubles to 3.88 pp, only 2.91 pp lower than {\system}.
Moreover, the fidelity of both comes at substantial profiling cost (see \S\ref{subsec:eval_profiling}).
Finally, the proxy-based baseline relies on public benchmark scores, failing to capture dataset-specific behavior and inter-stage dependencies, yielding the highest estimation error (20.48 pp median).

\begin{figure}[t]
\centering
\includegraphics[width=0.95\linewidth,]{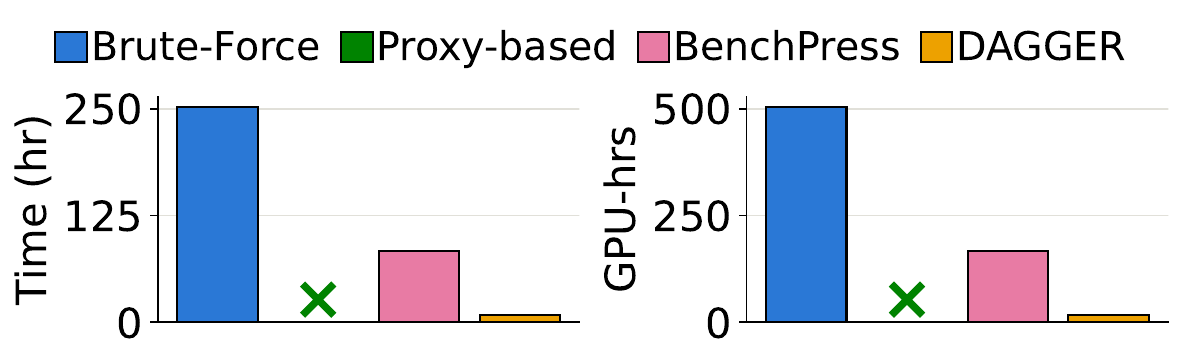}
\vspace{-4mm}
\caption{Profiling time and GPU-hours 
across baselines.
{\system} has minimal profiling overhead. See~\S~\ref{subsec:eval_profiling}.}
\label{fig:eval_profcost}
\vspace{-4mm}
\end{figure}

\vspace{-2mm}
\subsection{{\system}'s Profiling Cost}
\label{subsec:eval_profiling}
\vspace{-1mm}

We next evaluate the profiling overhead of {\system} against the baselines.
Figure~\ref{fig:eval_profcost} summarizes the profiling time and GPU-hours.
Brute-Force incurs substantial profiling overheads as it must profile all possible workflow configurations across all datasets, consuming $30.61\times$ more GPU-hours compared to {\system}.
BenchPress reduces this cost, as it only requires profiling a subset of those; however, it still requires $10.20\times$ more profiling time and GPU-hours compared to {\system}.
Proxy-based approaches incur negligible profiling cost, but suffer from poor estimation fidelity (see~\S~\ref{subsec:eval_accuracy}).

{\system} balances profiling cost and estimation fidelity by estimating per-node accuracy and per-edge dependency terms from sparse observations using SVD-based matrix completion.
These terms are then used to estimate workflow quality, allowing {\system} to closely match ground-truth quality with minimal profiling overhead.
{\system} also remains effective for previously unseen models and datasets, where only a small number of entries must be profiled before estimating the remaining values through matrix completion.

\vspace{-2mm}
\subsection{{\system}'s Workflow Selection Ability}
\label{subsec:eval_wfselection}
\vspace{-1mm}

To evaluate {\system}'s ability to identify the optimal workflow configuration, we fix the sequential workflow and, for each dataset, estimate the quality of all 27 candidate configurations, selecting the one with the highest estimated quality; complete results are reported in Table~\ref{tab:eval_selected_wfs} (Appendix~\ref{sec:appendix_a}).
{\system} selects the optimal configuration on 2WikiMultiHopQA, the second-best configuration on HotpotQA (0.5 pp below optimal),  and a median-ranked configuration on MuSiQue (14th of 27, 6.5 pp below optimal), where the top candidates exhibit tightly clustered quality values that are difficult to distinguish under estimation error. 
The proxy-based baseline invariably selects the same all-Qwen configuration, as this maximizes its dataset-agnostic proxy score. This choice happens to be optimal on MuSiQue but suboptimal on the other two datasets.
LLMSelector prunes the search space and does not explore all configurations, missing the optimum on every dataset: it picks the 4th, 5th, and 3rd ranked configuration for HotpotQA, MuSiQue, and 2WikiMultiHopQA, respectively.
BenchPress identifies the optimum on all three datasets, but only because it falls among its directly observed entries during matrix completion rather than among those it genuinely estimates. Overall, DAGGER consistently makes strong selections using only cost-efficient estimation without extended profiling. 

%% file: 05_Related.tex
\vspace{-2mm}
\section{Related Work}
\label{sec:related}
\vspace{-1mm}

\noindent\textbf{Quality Evaluation for Generative and Agentic AI. }
Prior work introduced benchmark suites for evaluating agentic systems across reasoning, coding, and tool-use tasks, including the Open LLM Leaderboard~\cite{openllmleaderboard}, SWE-Bench~\cite{swebench}, Terminal-Bench~\cite{terminalbench}, and GAIA~\cite{gaia}.
These measure end-to-end task success or judge-based preference scores using human evaluators or LLM-as-a-judge frameworks such as MT-Bench~\cite{mtbench}, AlpacaEval~\cite{alpacaeval}, and G-Eval~\cite{geval}.
However, they require executing instantiated models or workflows and thus cannot estimate the quality of unseen workflows before deployment.

\noindent\textbf{Workflow optimization.} Recent work explored model selection and execution optimization for compound and agentic AI systems.
LLMSelector~\cite{LLMSelector} selects models for compound AI modules, Aragog~\cite{Aragog} dynamically routes requests across workflow configurations, and Murakkab~\cite{Murakkab} profiles candidate workflows to estimate quality.
These systems reduce the number of evaluated configurations, but still require profiling models or workflows, which becomes expensive as workflow complexity and the number of candidate models increase.
{\system} instead estimates workflow quality analytically from per-node accuracy terms, per-edge dependency terms, and branch weights without deploying candidate workflows.

\noindent \textbf{Reducing profiling cost via matrix completion.}
Classical matrix-completion methods recover low-rank matrices from sparse observations~\cite{candes2009exact,candes2010power,keshavan2010matrix}, while practical spectral and SVD-based methods make such recovery scalable~\cite{cai2010singular}.
Recent work applied similar low-rank structure to LLM evaluation, including tensor-completion formulations for sparse leaderboard inference~\cite{li2026llmevaltensor, Papailiopoulos}.
Unlike these approaches, which estimate missing entries in model--benchmark tables or evaluation tensors, {\system} infers per-node accuracy terms and per-edge dependency terms from sparse workflow observations to estimate the end-to-end quality of unseen workflow DAGs.

%% file: 06_Conclusion.tex
\vspace{-2mm}
\section{Conclusion}
\label{sec:conclusion}
\vspace{-1mm}

We introduced {\system}, a framework for analytically estimating the end-to-end quality of generative AI workflows without exhaustive deployment and profiling.
{\system} models workflow execution over DAGs using per-node accuracy terms and per-edge dependency terms inferred from sparse observations via SVD-based matrix completion.
Our evaluation showed that {\system} accurately estimates workflow quality while substantially reducing profiling cost compared to state-of-the-art approaches.

%% file: 07_Appendix_A.tex
\begin{strip}
\begin{minipage}[t][0.90\textheight][t]{\textwidth}

\noindent
\begin{minipage}[t][0.245\textheight][t]{0.485\linewidth}
\vspace{0pt}

\section{Supplementary Material}
\label{sec:appendix_a}

\vspace{0.1cm}

\subsection{Matrix Completion Results}
\label{sec:appendix_matrix_completion}

\vspace{0.1cm}

The first step of {\system} applies SVD-based matrix completion to the per-node accuracy and per-edge dependency terms to minimize the associated profiling cost. 
Our evaluation setup consists of three models: $M_1$ (Qwen-2.5-7B), $M_2$ (Mistral-7B), and $M_3$ (Llama-3.1-8B); three datasets: $D_1$ (HotpotQA), $D_2$ (MuSiQue), and $D_3$ (2WikiMultiHopQA); and three workflows: $WF_1$ (sequential), $WF_2$ (branching), and $WF_3$ (arbitrary). 
Across the workflows, there are seven unique operations: (a)~Extract$_1$, (b)~Reason, (c)~Generate, (d)~Analyze, (e)~Extract$_2$, (f)~Link, and (g)~Synthesize (see Figure~\ref{fig:eval_wfs}). The results of the two SVD-based completion tasks are presented below.

\noindent\textbf{Node Accuracy.}
We randomly observe 53 of the 63 total entries and apply SVD-based matrix completion with a rank-1 approximation to estimate the remaining entries.
Table~\ref{tab:eval_pernode_ALL} shows the relative errors of the estimated values compared to ground truth for all models and dataset-specific operations, where $D_i^{k}$ denotes operation $k \in \{a,b,c,d,e,f,g\}$ executed on dataset $i \in \{1, 2, 3\}$. Observed values are represented with "$-$". The median estimation error is 19.1\%.

\noindent\textbf{Edge Dependency.}
Following the same methodology, we apply rank-1 SVD-based matrix completion
to the per-edge dependency terms. The key difference is that dependency values
are indexed by model combinations and dataset-specific edges rather than
individual models and operations. Table~\ref{tab:eval_peredge_ALL} presents
the estimation error for the complete matrix across all model combinations
and dataset-specific edges. We denote by $MC_j^i$ that model $M_i$ is connected to model $M_j$, and by $D_i^{kl}$ the edge from operation $k$ to
operation $l$ on dataset $i$, where $i \in \{1, 2, 3\}$ and
$k, l \in \{a, \ldots, g\}$. The median estimation error is 16.4\%.

\vfill
\end{minipage}%
\hfill
\begin{minipage}[t][0.245\textheight][t]{0.485\linewidth}
\vspace{0pt}

\subsection{{\system}'s Workflow Selection Results}
\label{sec:appendix_selection_results}

We compare {\system} against the baselines in selecting the optimal workflow configuration for a given workflow--dataset combination.
As all baselines except LLMSelector estimate the end-to-end quality of a given workflow configuration rather than directly performing selection, we use each baseline to estimate the quality of all candidate configurations and select the one with the highest estimated quality.
Table~\ref{tab:eval_selected_wfs} presents the results for the sequential workflow ($\mathrm{Extract} \rightarrow \mathrm{Reason} \rightarrow \mathrm{Generate}$) across all three datasets, reporting the model assigned to each operation, the rank of the selected configuration among the 27 candidates, its real quality, and the optimality gap relative to the best configuration.
Brute-Force selects the optimal configuration on all datasets, as it exhaustively profiles every candidate; however, it incurs substantial profiling costs.
The proxy-based baseline maximizes a dataset-agnostic proxy score and therefore invariably selects the all-Qwen configuration; this choice happens to be optimal on MuSiQue but is suboptimal on HotpotQA ($-5.0$~pp) and 2WikiMultiHopQA ($-11.0$~pp).
BenchPress also identifies the optimal configuration on all three datasets; however, in each case the optimum falls among the entries it observes directly during matrix completion, and thus its selection does not reflect genuine estimation.
LLMSelector's data-driven approach does not explore the full configuration space and thus never selects the optimal configuration; it selects the 4th-ranked configuration on HotpotQA ($-1.0$~pp), the 5th on MuSiQue ($-3.0$~pp), and the 3rd on 2WikiMultiHopQA ($-0.5$~pp). 
Finally, {\system} selects the second-best configuration on HotpotQA ($-0.5$~pp) and the optimal configuration on 2WikiMultiHopQA. On MuSiQue, it selects a lower-ranked configuration (14th of 27, $-6.5$~pp), as the top-ranked configurations exhibit tightly clustered quality values that are difficult to distinguish under estimation error.

\vfill
\end{minipage}

\vspace{8cm}

\begin{minipage}{\textwidth}
\centering
\small
\setlength{\tabcolsep}{2pt}
\renewcommand{\arraystretch}{1.2}

\captionof{table}{Per-node accuracy values across all three models and 21 dataset-specific operations. Each entry reports the ground-truth/completed value, followed by the relative percentage estimation error in parentheses. A dash indicates an observed entry.}
\label{tab:eval_pernode_ALL}

\begin{tabularx}{\linewidth}{l||*{7}{Y}}
\toprule
& \multicolumn{7}{c}{$\boldsymbol{D_1:\text{HotpotQA}}$} \\
& $\boldsymbol{D_1^a}$ & $\boldsymbol{D_1^b}$ & $\boldsymbol{D_1^c}$ & $\boldsymbol{D_1^d}$ & $\boldsymbol{D_1^e}$ & $\boldsymbol{D_1^f}$ & $\boldsymbol{D_1^g}$ \\
\midrule
$\boldsymbol{M_1}$ & $0.84/0.84\ (-)$ & $0.32/0.32\ (-)$ & $0.82/0.82\ (-)$ & $0.85/0.85\ (-)$ & $0.57/0.57\ (-)$ & $0.34/0.34\ (-)$ & $0.33/0.43\ (28.8)$ \\
$\boldsymbol{M_2}$ & $0.69/0.69\ (-)$ & $0.51/0.51\ (-)$ & $0.66/0.66\ (-)$ & $0.85/0.59\ (30.3)$ & $0.71/0.71\ (-)$ & $0.33/0.33\ (-)$ & $0.52/0.43\ (17.5)$ \\
$\boldsymbol{M_3}$ & $0.73/0.73\ (-)$ & $0.49/0.49\ (-)$ & $0.75/0.75\ (-)$ & $0.51/0.51\ (-)$ & $0.69/0.69\ (-)$ & $0.61/0.61\ (-)$ & $0.43/0.43\ (-)$ \\
\bottomrule
\end{tabularx}

\par\vspace{1em}

\begin{tabularx}{\linewidth}{l||*{7}{Y}}
\toprule
& \multicolumn{7}{c}{$\boldsymbol{D_2:\text{MuSiQue}}$} \\
& $\boldsymbol{D_2^a}$ & $\boldsymbol{D_2^b}$ & $\boldsymbol{D_2^c}$ & $\boldsymbol{D_2^d}$ & $\boldsymbol{D_2^e}$ & $\boldsymbol{D_2^f}$ & $\boldsymbol{D_2^g}$ \\
\midrule
$\boldsymbol{M_1}$ & $0.55/0.55\ (-)$ & $0.26/0.26\ (-)$ & $0.52/0.52\ (-)$ & $0.51/0.40\ (20.8)$ & $0.27/0.27\ (-)$ & $0.12/0.12\ (-)$ & $0.23/0.23\ (-)$ \\
$\boldsymbol{M_2}$ & $0.42/0.42\ (-)$ & $0.30/0.30\ (-)$ & $0.43/0.43\ (-)$ & $0.55/0.40\ (26.6)$ & $0.38/0.38\ (-)$ & $0.22/0.22\ (-)$ & $0.33/0.23\ (29.6)$ \\
$\boldsymbol{M_3}$ & $0.48/0.48\ (-)$ & $0.24/0.24\ (-)$ & $0.52/0.44\ (14.8)$ & $0.40/0.40\ (-)$ & $0.36/0.36\ (2.7)$ & $0.39/0.39\ (-)$ & $0.23/0.23\ (-)$ \\
\bottomrule
\end{tabularx}

\par\vspace{1em}

\begin{tabularx}{\linewidth}{l||*{7}{Y}}
\toprule
& \multicolumn{7}{c}{$\boldsymbol{D_3:\text{2WikiMultiHopQA}}$} \\
& $\boldsymbol{D_3^a}$ & $\boldsymbol{D_3^b}$ & $\boldsymbol{D_3^c}$ & $\boldsymbol{D_3^d}$ & $\boldsymbol{D_3^e}$ & $\boldsymbol{D_3^f}$ & $\boldsymbol{D_3^g}$ \\
\midrule
$\boldsymbol{M_1}$ & $0.80/0.73\ (9.1)$ & $0.26/0.26\ (-)$ & $0.71/0.71\ (-)$ & $0.77/0.77\ (1.1)$ & $0.48/0.48\ (-)$ & $0.19/0.19\ (-)$ & $0.24/0.24\ (-)$ \\
$\boldsymbol{M_2}$ & $0.71/0.71\ (-)$ & $0.24/0.24\ (-)$ & $0.50/0.50\ (-)$ & $0.89/0.89\ (-)$ & $0.65/0.65\ (-)$ & $0.31/0.31\ (-)$ & $0.26/0.26\ (-)$ \\
$\boldsymbol{M_3}$ & $0.76/0.76\ (-)$ & $0.24/0.24\ (-)$ & $0.69/0.69\ (-)$ & $0.56/0.56\ (-)$ & $0.76/0.76\ (-)$ & $0.55/0.55\ (-)$ & $0.27/0.27\ (-)$ \\
\bottomrule
\end{tabularx}

\end{minipage}
\end{minipage}

\end{strip}

\clearpage
\noindent
\makebox[\textwidth][c]{%
\begin{minipage}{1.1\textwidth}
\centering
\small
\setlength{\tabcolsep}{2pt}
\renewcommand{\arraystretch}{1.2}

\captionof{table}{Per-edge dependency values across all 9 model combinations and 27 dataset-specific edges. Each entry reports the ground-truth/completed value, followed by the relative percentage estimation error in parentheses. A dash indicates an observed entry.}
\label{tab:eval_peredge_ALL}

\begin{tabularx}{\linewidth}{l||*{9}{Y}}
\toprule
& \multicolumn{9}{c}{$\boldsymbol{D_1:\mathrm{HotpotQA}}$} \\
& $\boldsymbol{D_1^{ab}}$ & $\boldsymbol{D_1^{bc}}$ & $\boldsymbol{D_1^{de}}$ & $\boldsymbol{D_1^{df}}$ & $\boldsymbol{D_1^{ec}}$ & $\boldsymbol{D_1^{fc}}$ & $\boldsymbol{D_1^{eg}}$ & $\boldsymbol{D_1^{fg}}$ & $\boldsymbol{D_1^{gc}}$ \\
\midrule
$\boldsymbol{MC_1^1}$ & $1.0/1.0\ (-)$ & $2.9/2.9\ (-)$ & $1.2/1.6\ (29.9)$ & $1.1/1.1\ (-)$ & $1.5/1.5\ (1.6)$ & $2.5/2.4\ (4.8)$ & $1.5/1.4\ (9.8)$ & $2.4/2.7\ (8.9)$ & $2.9/3.1\ (7.2)$ \\
$\boldsymbol{MC_2^1}$ & $1.0/1.0\ (4.8)$ & $3.5/3.5\ (-)$ & $1.2/1.6\ (33.4)$ & $1.5/1.1\ (24.6)$ & $1.6/1.6\ (-)$ & $2.8/2.8\ (-)$ & $1.4/1.4\ (-)$ & $2.5/2.9\ (16.4)$ & $3.4/3.4\ (-)$ \\
$\boldsymbol{MC_3^1}$ & $1.0/1.0\ (-)$ & $3.0/3.3\ (11.3)$ & $1.4/1.4\ (-)$ & $1.3/1.3\ (-)$ & $1.6/1.6\ (-)$ & $2.7/2.6\ (6.3)$ & $1.4/1.4\ (-)$ & $2.5/2.8\ (9.7)$ & $3.2/3.2\ (-)$ \\
$\boldsymbol{MC_1^2}$ & $1.0/1.1\ (5.6)$ & $1.9/1.9\ (-)$ & $1.1/1.4\ (27.4)$ & $0.4/1.4\ (225.6)$ & $1.2/1.2\ (-)$ & $2.1/2.1\ (-)$ & $1.1/1.1\ (-)$ & $2.1/2.1\ (-)$ & $1.8/2.3\ (26.4)$ \\
$\boldsymbol{MC_2^2}$ & $1.0/1.0\ (-)$ & $2.1/2.2\ (7.0)$ & $1.1/1.1\ (-)$ & $1.5/1.5\ (-)$ & $1.4/1.4\ (-)$ & $2.4/1.7\ (28.9)$ & $1.3/1.3\ (-)$ & $1.9/2.1\ (11.5)$ & $1.9/1.9\ (-)$ \\
$\boldsymbol{MC_3^2}$ & $1.1/1.1\ (-)$ & $2.0/2.4\ (24.2)$ & $1.4/1.4\ (-)$ & $1.2/1.2\ (-)$ & $1.3/1.3\ (-)$ & $2.4/2.4\ (-)$ & $1.3/1.3\ (-)$ & $2.3/2.3\ (-)$ & $1.8/2.5\ (40.1)$ \\
$\boldsymbol{MC_1^3}$ & $1.2/1.2\ (-)$ & $2.0/2.1\ (7.1)$ & $1.5/1.5\ (-)$ & $1.3/1.3\ (-)$ & $1.3/1.3\ (-)$ & $1.3/1.3\ (-)$ & $1.0/1.0\ (-)$ & $1.5/2.0\ (35.0)$ & $2.2/2.2\ (-)$ \\
$\boldsymbol{MC_2^3}$ & $1.2/1.1\ (8.6)$ & $2.3/2.3\ (-)$ & $1.6/1.5\ (10.0)$ & $1.7/1.3\ (23.9)$ & $1.3/1.3\ (-)$ & $1.5/1.5\ (-)$ & $1.5/1.5\ (-)$ & $1.9/2.2\ (19.6)$ & $2.7/2.7\ (-)$ \\
$\boldsymbol{MC_3^3}$ & $1.1/1.1\ (-)$ & $2.1/2.7\ (28.1)$ & $2.0/2.0\ (-)$ & $1.8/1.3\ (26.8)$ & $1.2/1.2\ (-)$ & $1.5/2.1\ (40.9)$ & $1.8/1.3\ (27.3)$ & $2.7/2.7\ (-)$ & $2.4/2.7\ (16.5)$ \\
\bottomrule
\end{tabularx}

\par\vspace{1.2em}

\begin{tabularx}{\linewidth}{l||*{9}{Y}}
\toprule
& \multicolumn{9}{c}{$\boldsymbol{D_2:\mathrm{MuSiQue}}$} \\
& $\boldsymbol{D_2^{ab}}$ & $\boldsymbol{D_2^{bc}}$ & $\boldsymbol{D_2^{de}}$ & $\boldsymbol{D_2^{df}}$ & $\boldsymbol{D_2^{ec}}$ & $\boldsymbol{D_2^{fc}}$ & $\boldsymbol{D_2^{eg}}$ & $\boldsymbol{D_2^{fg}}$ & $\boldsymbol{D_2^{gc}}$ \\
\midrule
$\boldsymbol{MC_1^1}$ & $0.9/0.9\ (-)$ & $3.6/3.6\ (-)$ & $2.7/2.7\ (1.4)$ & $2.7/2.7\ (-)$ & $3.0/3.0\ (-)$ & $5.5/4.4\ (21.2)$ & $2.4/2.4\ (-)$ & $6.1/6.1\ (-)$ & $4.2/3.9\ (8.8)$ \\
$\boldsymbol{MC_2^1}$ & $1.2/1.2\ (-)$ & $4.1/4.1\ (-)$ & $3.0/3.0\ (-)$ & $1.9/2.7\ (45.4)$ & $2.9/3.2\ (11.2)$ & $5.7/5.7\ (-)$ & $2.8/2.6\ (9.3)$ & $5.6/7.2\ (28.4)$ & $4.9/4.4\ (8.5)$ \\
$\boldsymbol{MC_3^1}$ & $1.4/1.4\ (-)$ & $3.4/3.4\ (-)$ & $3.5/2.9\ (19.3)$ & $2.2/2.2\ (-)$ & $3.0/3.0\ (0.0)$ & $5.5/5.5\ (-)$ & $3.0/2.5\ (17.4)$ & $6.2/6.2\ (-)$ & $4.1/4.1\ (-)$ \\
$\boldsymbol{MC_1^2}$ & $0.7/0.7\ (-)$ & $3.1/3.7\ (19.5)$ & $2.2/2.2\ (-)$ & $1.3/1.3\ (-)$ & $2.2/2.2\ (-)$ & $2.7/2.7\ (-)$ & $2.4/2.1\ (12.4)$ & $2.7/2.8\ (5.8)$ & $2.9/2.5\ (13.4)$ \\
$\boldsymbol{MC_2^2}$ & $1.2/1.2\ (1.0)$ & $3.1/3.1\ (-)$ & $1.9/2.1\ (10.7)$ & $1.9/1.6\ (12.9)$ & $2.2/2.0\ (12.3)$ & $2.1/2.1\ (-)$ & $2.1/2.1\ (-)$ & $3.1/3.1\ (-)$ & $2.7/2.7\ (-)$ \\
$\boldsymbol{MC_3^2}$ & $1.3/1.2\ (10.2)$ & $2.8/2.8\ (-)$ & $2.8/2.3\ (19.4)$ & $1.6/1.6\ (-)$ & $2.1/2.1\ (-)$ & $2.6/2.6\ (-)$ & $2.7/2.2\ (18.7)$ & $3.6/3.6\ (-)$ & $2.6/2.6\ (-)$ \\
$\boldsymbol{MC_1^3}$ & $1.1/1.1\ (1.1)$ & $4.2/3.7\ (12.2)$ & $1.8/1.8\ (-)$ & $1.8/1.5\ (14.8)$ & $2.3/1.8\ (21.8)$ & $1.8/1.8\ (-)$ & $2.0/2.1\ (4.8)$ & $2.1/2.6\ (19.9)$ & $4.6/2.4\ (48.3)$ \\
$\boldsymbol{MC_2^3}$ & $1.5/1.5\ (-)$ & $5.0/5.0\ (-)$ & $2.2/2.2\ (-)$ & $2.1/2.1\ (-)$ & $2.0/2.0\ (-)$ & $1.9/1.9\ (-)$ & $3.0/2.2\ (26.0)$ & $2.7/2.7\ (-)$ & $5.2/2.8\ (44.8)$ \\
$\boldsymbol{MC_3^3}$ & $1.5/1.5\ (-)$ & $4.2/4.2\ (-)$ & $2.7/2.7\ (-)$ & $2.0/2.0\ (-)$ & $2.1/2.4\ (16.0)$ & $1.8/1.8\ (-)$ & $4.3/2.3\ (47.1)$ & $5.1/4.5\ (11.5)$ & $4.6/3.2\ (30.6)$ \\
\bottomrule
\end{tabularx}

\par\vspace{1.2em}

\begin{tabularx}{\linewidth}{l||*{9}{Y}}
\toprule
& \multicolumn{9}{c}{$\boldsymbol{D_3:\mathrm{2WikiMultiHopQA}}$} \\
& $\boldsymbol{D_3^{ab}}$ & $\boldsymbol{D_3^{bc}}$ & $\boldsymbol{D_3^{de}}$ & $\boldsymbol{D_3^{df}}$ & $\boldsymbol{D_3^{ec}}$ & $\boldsymbol{D_3^{fc}}$ & $\boldsymbol{D_3^{eg}}$ & $\boldsymbol{D_3^{fg}}$ & $\boldsymbol{D_3^{gc}}$ \\
\midrule
$\boldsymbol{MC_1^1}$ & $0.7/1.2\ (76.0)$ & $3.3/4.2\ (24.2)$ & $1.5/1.5\ (-)$ & $1.1/1.1\ (-)$ & $1.8/1.8\ (-)$ & $2.9/2.6\ (11.6)$ & $1.7/1.7\ (-)$ & $2.5/2.5\ (-)$ & $3.7/3.7\ (-)$ \\
$\boldsymbol{MC_2^1}$ & $1.2/1.2\ (-)$ & $4.6/4.6\ (-)$ & $1.3/1.3\ (-)$ & $1.5/1.1\ (25.2)$ & $1.9/2.1\ (11.2)$ & $3.5/3.5\ (-)$ & $2.3/2.3\ (-)$ & $3.3/3.3\ (1.8)$ & $4.9/4.0\ (17.9)$ \\
$\boldsymbol{MC_3^1}$ & $1.0/1.2\ (26.7)$ & $3.6/3.6\ (-)$ & $1.5/1.5\ (-)$ & $1.2/1.2\ (-)$ & $1.9/1.9\ (-)$ & $2.8/2.9\ (2.0)$ & $2.5/2.5\ (-)$ & $3.4/3.4\ (-)$ & $3.9/3.9\ (-)$ \\
$\boldsymbol{MC_1^2}$ & $0.7/0.9\ (33.6)$ & $3.8/3.8\ (-)$ & $1.4/1.4\ (-)$ & $1.1/1.4\ (36.1)$ & $1.2/1.2\ (0.1)$ & $1.7/1.3\ (25.9)$ & $1.0/1.9\ (98.8)$ & $1.3/1.3\ (-)$ & $3.0/3.0\ (1.1)$ \\
$\boldsymbol{MC_2^2}$ & $0.8/0.8\ (-)$ & $3.8/4.1\ (6.8)$ & $1.0/1.0\ (-)$ & $1.5/1.5\ (-)$ & $1.1/1.2\ (4.8)$ & $1.8/1.8\ (-)$ & $1.5/1.9\ (26.3)$ & $2.3/2.3\ (-)$ & $3.1/3.1\ (-)$ \\
$\boldsymbol{MC_3^2}$ & $0.8/0.8\ (-)$ & $3.6/3.6\ (-)$ & $1.3/1.3\ (-)$ & $1.1/1.1\ (-)$ & $1.2/1.2\ (-)$ & $1.8/1.6\ (10.5)$ & $1.4/1.4\ (-)$ & $2.5/2.5\ (-)$ & $2.9/2.9\ (-)$ \\
$\boldsymbol{MC_1^3}$ & $0.9/0.9\ (2.6)$ & $3.9/3.9\ (-)$ & $1.4/1.4\ (-)$ & $1.6/1.6\ (-)$ & $1.1/1.1\ (2.6)$ & $1.4/1.4\ (-)$ & $0.8/1.9\ (139.2)$ & $2.0/2.0\ (-)$ & $3.2/3.2\ (-)$ \\
$\boldsymbol{MC_2^3}$ & $1.4/1.4\ (-)$ & $5.3/5.3\ (-)$ & $1.4/1.3\ (3.0)$ & $1.4/1.4\ (-)$ & $1.1/1.1\ (-)$ & $1.4/1.4\ (-)$ & $2.2/2.2\ (-)$ & $3.2/2.1\ (33.7)$ & $4.0/3.2\ (20.4)$ \\
$\boldsymbol{MC_3^3}$ & $1.0/1.0\ (-)$ & $4.0/4.0\ (-)$ & $1.4/1.3\ (5.3)$ & $1.0/1.3\ (30.1)$ & $1.0/1.5\ (46.7)$ & $1.5/1.5\ (-)$ & $2.2/2.2\ (-)$ & $3.8/2.4\ (37.0)$ & $3.4/3.4\ (-)$ \\
\bottomrule
\end{tabularx}

\par\vspace{2.5em}

\captionof{table}{Selected workflow configuration per baseline for each dataset using the sequential workflow. Each row reports the model assigned to each operation, the rank among all 27 candidates, the real quality, and the optimality gap in percentage points (pp). Configurations that are observed are denoted with $\dagger$.}
\label{tab:eval_selected_wfs}

\begin{tabularx}{\linewidth}{l||YYY|Y|Y|Y}
\toprule
& \multicolumn{6}{c}{$\boldsymbol{\mathrm{HotpotQA}}$} \\
\textbf{Baseline} & \textbf{Extractor} & \textbf{Reasoner} & \textbf{Answer Gen.} & \textbf{Rank} & \textbf{Real Quality (\%)} & \textbf{Gap (pp)} \\
\midrule
\textbf{Brute-Force} & Qwen-2.5-7B & Llama-3.1-8B & Qwen-2.5-7B  & 1/27  & 66.5 & 0.0 $\dagger$ \\
\textbf{Proxy-based} & Qwen-2.5-7B & Qwen-2.5-7B  & Qwen-2.5-7B  & 11/27 & 61.5 & $-5.0$ \\
\textbf{BenchPress}  & Qwen-2.5-7B & Llama-3.1-8B & Qwen-2.5-7B  & 1/27  & 66.5 & 0.0 $\dagger$ \\
\textbf{LLMSelector} & Qwen-2.5-7B & Mistral-7B   & Qwen-2.5-7B  & 4/27  & 65.5 & $-1.0$ \\
\textbf{{\system}}   & Qwen-2.5-7B & Llama-3.1-8B & Llama-3.1-8B & 2/27  & 66.0 & $-0.5$ \\

\bottomrule
\end{tabularx}

\end{minipage}%
}

\clearpage
\noindent
\makebox[\textwidth][c]{%
\begin{minipage}{1.1\textwidth}
\centering
\small
\setlength{\tabcolsep}{4pt}
\renewcommand{\arraystretch}{1.2}

\begin{tabularx}{\linewidth}{l||YYY|Y|Y|Y}
\toprule
& \multicolumn{6}{c}{$\boldsymbol{\mathrm{MuSiQue}}$} \\
\textbf{Baseline} & \textbf{Extractor} & \textbf{Reasoner} & \textbf{Answer Gen.} & \textbf{Rank} & \textbf{Real Quality (\%)} & \textbf{Gap (pp)} \\
\midrule
\textbf{Brute-Force} & Qwen-2.5-7B  & Qwen-2.5-7B  & Qwen-2.5-7B  & 1/27  & 35.5 & $0.0$ $\dagger$ \\
\textbf{Proxy-based} & Qwen-2.5-7B  & Qwen-2.5-7B  & Qwen-2.5-7B  & 1/27  & 35.5 & $0.0$ \\
\textbf{BenchPress}  & Qwen-2.5-7B  & Qwen-2.5-7B  & Qwen-2.5-7B  & 1/27  & 35.5 & $0.0$ $\dagger$ \\
\textbf{LLMSelector} & Qwen-2.5-7B  & Llama-3.1-8B & Llama-3.1-8B & 5/27  & 32.5 & $-3.0$ \\
\textbf{{\system}}   & Llama-3.1-8B & Mistral-7B   & Qwen-2.5-7B  & 14/27 & 29.0 & $-6.5$ \\
\bottomrule
\end{tabularx}

\par\vspace{1.2em}

\begin{tabularx}{\linewidth}{l||YYY|Y|Y|Y}
\toprule
& \multicolumn{6}{c}{$\boldsymbol{\mathrm{2WikiMultiHopQA}}$} \\
\textbf{Baseline} & \textbf{Extractor} & \textbf{Reasoner} & \textbf{Answer Gen.} & \textbf{Rank} & \textbf{Real Quality (\%)} & \textbf{Gap (pp)} \\
\midrule
\textbf{Brute-Force} & Llama-3.1-8B & Mistral-7B   & Qwen-2.5-7B  & 1/27  & 55.0 & $0.0$ $\dagger$ \\
\textbf{Proxy-based} & Qwen-2.5-7B  & Qwen-2.5-7B  & Qwen-2.5-7B  & 15/27 & 44.0 & $-11.0$ \\
\textbf{BenchPress}  & Llama-3.1-8B & Mistral-7B   & Qwen-2.5-7B  & 1/27  & 55.0 & $0.0$ $\dagger$ \\
\textbf{LLMSelector} & Qwen-2.5-7B  & Llama-3.1-8B & Llama-3.1-8B & 3/27  & 54.5 & $-0.5$ \\
\textbf{{\system}}   & Llama-3.1-8B & Mistral-7B   & Qwen-2.5-7B  & 1/27  & 55.0 & $0.0$ \\
\bottomrule
\end{tabularx}

\end{minipage}%
}

%% file: refs.bib
@misc{BAIR_CompoundAI,
  title={{The Shift from Models to Compound AI Systems}},
  author={Matei Zaharia and Omar Khattab and Lingjiao Chen and Jared Quincy Davis
          and Heather Miller and Chris Potts and James Zou and Michael Carbin
          and Jonathan Frankle and Naveen Rao and Ali Ghodsi},
  howpublished={\url{https://bair.berkeley.edu/blog/2024/02/18/compound-ai-systems/}},
  year={2024}
}

@misc{Murakkab,
      title={{Murakkab: Resource-Efficient Agentic Workflow Orchestration in Cloud Platforms}}, 
      author={Gohar Irfan Chaudhry and Esha Choukse and Haoran Qiu and Íñigo Goiri and Rodrigo Fonseca and Adam Belay and Ricardo Bianchini},
      year={2025},
      eprint={2508.18298},
      archivePrefix={arXiv},
      primaryClass={cs.MA},
      url={https://arxiv.org/abs/2508.18298}, 
}

@misc{Aragog,
      title={Aragog: Just-in-Time Model Routing for Scalable Serving of Agentic Workflows}, 
      author={Yinwei Dai and Zhuofu Chen and Anand Iyer and Ravi Netravali},
      year={2025},
      eprint={2511.20975},
      archivePrefix={arXiv},
      primaryClass={cs.DC},
      url={https://arxiv.org/abs/2511.20975}, 
}

@misc{Papailiopoulos,
      title={You Don't Need to Run Every Eval}, 
      author={Yuchen Zeng and Dimitris Papailiopoulos},
      year={2026},
      eprint={2606.24020},
      archivePrefix={arXiv},
      primaryClass={cs.LG},
      url={https://arxiv.org/abs/2606.24020}, 
}

@inproceedings{hotpotqa,
    title = "{{H}otpot{QA}: A Dataset for Diverse, Explainable Multi-hop Question Answering}",
    author = "Yang, Zhilin  and
      Qi, Peng  and
      Zhang, Saizheng  and
      Bengio, Yoshua  and
      Cohen, William  and
      Salakhutdinov, Ruslan  and
      Manning, Christopher D.",
    editor = "Riloff, Ellen  and
      Chiang, David  and
      Hockenmaier, Julia  and
      Tsujii, Jun{'}ichi",
    booktitle = "Proceedings of the 2018 Conference on Empirical Methods in Natural Language Processing",
    month = oct # "-" # nov,
    year = "2018",
    address = "Brussels, Belgium",
    publisher = "Association for Computational Linguistics",
    url = "https://aclanthology.org/D18-1259/",
    doi = "10.18653/v1/D18-1259",
    pages = "2369--2380"
}

@misc{qwen-2.5,
      title={{Qwen2.5 Technical Report}}, 
      author={Qwen and : and An Yang and Baosong Yang and Beichen Zhang and Binyuan Hui and Bo Zheng and Bowen Yu and Chengyuan Li and Dayiheng Liu and Fei Huang and Haoran Wei and Huan Lin and Jian Yang and Jianhong Tu and Jianwei Zhang and Jianxin Yang and Jiaxi Yang and Jingren Zhou and Junyang Lin and Kai Dang and Keming Lu and Keqin Bao and Kexin Yang and Le Yu and Mei Li and Mingfeng Xue and Pei Zhang and Qin Zhu and Rui Men and Runji Lin and Tianhao Li and Tianyi Tang and Tingyu Xia and Xingzhang Ren and Xuancheng Ren and Yang Fan and Yang Su and Yichang Zhang and Yu Wan and Yuqiong Liu and Zeyu Cui and Zhenru Zhang and Zihan Qiu},
      year={2025},
      eprint={2412.15115},
      archivePrefix={arXiv},
      primaryClass={cs.CL},
      url={https://arxiv.org/abs/2412.15115}, 
}

@article{llama-3,
  title={{The llama 3 herd of models}},
  author={Grattafiori, Aaron and Dubey, Abhimanyu and Jauhri, Abhinav and Pandey, Abhinav and Kadian, Abhishek and Al-Dahle, Ahmad and Letman, Aiesha and Mathur, Akhil and Schelten, Alan and Vaughan, Alex and others},
  journal={arXiv preprint arXiv:2407.21783},
  year={2024},
  url={https://arxiv.org/abs/2407.21783}
}

@misc{mistral,
      title={{Mistral 7B}}, 
      author={Albert Q. Jiang and Alexandre Sablayrolles and Arthur Mensch and Chris Bamford and Devendra Singh Chaplot and Diego de las Casas and Florian Bressand and Gianna Lengyel and Guillaume Lample and Lucile Saulnier and Lélio Renard Lavaud and Marie-Anne Lachaux and Pierre Stock and Teven Le Scao and Thibaut Lavril and Thomas Wang and Timothée Lacroix and William El Sayed},
      year={2023},
      eprint={2310.06825},
      archivePrefix={arXiv},
      primaryClass={cs.CL},
      url={https://arxiv.org/abs/2310.06825}, 
}

@misc{HuggingFace,
  author       = {{Hugging Face, Inc.}},
  title        = {{Hugging Face: The AI community building the future}},
  year         = {2026},
  howpublished = {\url{https://huggingface.co/}},
}

@misc{Accelerate,
    title = {Accelerate: Training and inference at scale made simple, efficient and adaptable.},
    author = {Sylvain Gugger and Lysandre Debut and Thomas Wolf and Philipp Schmid and Zachary Mueller and Sourab Mangrulkar and Marc Sun and Benjamin Bossan},
    howpublished = {\url{https://github.com/huggingface/accelerate}},
    year = {2022},
}

@article{FSDP,
    title={{PyTorch FSDP: Experiences on Scaling Fully Sharded Data Parallel}},
    author={Zhao, Yanli and Gu, Andrew and Varma, Rohan and Luo, Liang and Huang, Chien-Chin and Xu, Min and Wright, Less and Shojanazeri, Hamid and Ott, Myle and Shleifer, Sam and Desmaison, Alban and Balioglu, Can and Damania, Pritam and Nguyen, Bernard and Chauhan, Geeta and Hao, Yuchen and Mathews, Ajit and Li, Shen},
    journal={arXiv preprint arXiv:2304.11277},
    year={2023}
}

@misc{openai_gpt41_nano,
  author       = {{OpenAI}},
  title        = {{GPT-4.1 nano Model}},
  howpublished = {\url{https://developers.openai.com/api/docs/models/gpt-4.1-nano}},
  year         = {2025}
}

@inproceedings{MaverIQ,
author = {Liakopoulos, Dimitrios and Sinha, Prasoon and Hu, Tianrui and Lee, Myungjin and Yadwadkar, Neeraja J.},
title = {{MaverIQ: Fingerprint-Guided Extrapolation and Fragmentation-Aware Layering for Intent-Based LLM Serving}},
year = {2025},
isbn = {9798400714665},
publisher = {Association for Computing Machinery},
address = {New York, NY, USA},
url = {https://doi.org/10.1145/3712285.3759867},
doi = {10.1145/3712285.3759867},
booktitle = {Proceedings of the International Conference for High Performance Computing, Networking, Storage and Analysis},
pages = {1676–1696},
numpages = {21},
location = {
},
series = {SC '25}
}

@misc{openllmleaderboard,
  title        = {{Open LLM Leaderboard}},
  author       = {{Hugging Face H4 Team}},
  howpublished = {\url{https://huggingface.co/spaces/open-llm-leaderboard/}},
  year = 2024
}

@inproceedings{
swebench,
title={{SWE}-bench: Can Language Models Resolve Real-world Github Issues?},
author={Carlos E Jimenez and John Yang and Alexander Wettig and Shunyu Yao and Kexin Pei and Ofir Press and Karthik R Narasimhan},
booktitle={The Twelfth International Conference on Learning Representations},
year={2024},
url={https://openreview.net/forum?id=VTF8yNQM66}
}

@misc{terminalbench,
      title={Terminal-Bench: Benchmarking Agents on Hard, Realistic Tasks in Command Line Interfaces}, 
      author={Mike A. Merrill and Alexander G. Shaw and Nicholas Carlini and Boxuan Li and Harsh Raj and Ivan Bercovich and Lin Shi and Jeong Yeon Shin and Thomas Walshe and E. Kelly Buchanan and Junhong Shen and Guanghao Ye and Haowei Lin and Jason Poulos and Maoyu Wang and Marianna Nezhurina and Jenia Jitsev and Di Lu and Orfeas Menis Mastromichalakis and Zhiwei Xu and Zizhao Chen and Yue Liu and Robert Zhang and Leon Liangyu Chen and Anurag Kashyap and Jan-Lucas Uslu and Jeffrey Li and Jianbo Wu and Minghao Yan and Song Bian and Vedang Sharma and Ke Sun and Steven Dillmann and Akshay Anand and Andrew Lanpouthakoun and Bardia Koopah and Changran Hu and Etash Guha and Gabriel H. S. Dreiman and Jiacheng Zhu and Karl Krauth and Li Zhong and Niklas Muennighoff and Robert Amanfu and Shangyin Tan and Shreyas Pimpalgaonkar and Tushar Aggarwal and Xiangning Lin and Xin Lan and Xuandong Zhao and Yiqing Liang and Yuanli Wang and Zilong Wang and Changzhi Zhou and David Heineman and Hange Liu and Harsh Trivedi and John Yang and Junhong Lin and Manish Shetty and Michael Yang and Nabil Omi and Negin Raoof and Shanda Li and Terry Yue Zhuo and Wuwei Lin and Yiwei Dai and Yuxin Wang and Wenhao Chai and Shang Zhou and Dariush Wahdany and Ziyu She and Jiaming Hu and Zhikang Dong and Yuxuan Zhu and Sasha Cui and Ahson Saiyed and Arinbjörn Kolbeinsson and Jesse Hu and Christopher Michael Rytting and Ryan Marten and Yixin Wang and Alex Dimakis and Andy Konwinski and Ludwig Schmidt},
      year={2026},
      eprint={2601.11868},
      archivePrefix={arXiv},
      primaryClass={cs.SE},
      url={https://arxiv.org/abs/2601.11868}, 
}

@inproceedings{
gaia,
title={{GAIA}: a benchmark for General {AI} Assistants},
author={Gr{\'e}goire Mialon and Cl{\'e}mentine Fourrier and Thomas Wolf and Yann LeCun and Thomas Scialom},
booktitle={The Twelfth International Conference on Learning Representations},
year={2024},
url={https://openreview.net/forum?id=fibxvahvs3}
}

@inproceedings{mtbench,
author = {Zheng, Lianmin and Chiang, Wei-Lin and Sheng, Ying and Zhuang, Siyuan and Wu, Zhanghao and Zhuang, Yonghao and Lin, Zi and Li, Zhuohan and Li, Dacheng and Xing, Eric P. and Zhang, Hao and Gonzalez, Joseph E. and Stoica, Ion},
title = {Judging LLM-as-a-judge with MT-bench and Chatbot Arena},
year = {2023},
publisher = {Curran Associates Inc.},
address = {Red Hook, NY, USA},
booktitle = {Proceedings of the 37th International Conference on Neural Information Processing Systems},
articleno = {2020},
numpages = {29},
location = {New Orleans, LA, USA},
series = {NIPS '23}
}

@misc{alpacaeval,
      title={Length-Controlled AlpacaEval: A Simple Way to Debias Automatic Evaluators}, 
      author={Yann Dubois and Balázs Galambosi and Percy Liang and Tatsunori B. Hashimoto},
      year={2025},
      eprint={2404.04475},
      archivePrefix={arXiv},
      primaryClass={cs.LG},
      url={https://arxiv.org/abs/2404.04475}, 
}

@inproceedings{geval,
    title = "{G}-Eval: {NLG} Evaluation using Gpt-4 with Better Human Alignment",
    author = "Liu, Yang  and
      Iter, Dan  and
      Xu, Yichong  and
      Wang, Shuohang  and
      Xu, Ruochen  and
      Zhu, Chenguang",
    editor = "Bouamor, Houda  and
      Pino, Juan  and
      Bali, Kalika",
    booktitle = "Proceedings of the 2023 Conference on Empirical Methods in Natural Language Processing",
    month = dec,
    year = "2023",
    address = "Singapore",
    publisher = "Association for Computational Linguistics",
    url = "https://aclanthology.org/2023.emnlp-main.153/",
    doi = "10.18653/v1/2023.emnlp-main.153",
    pages = "2511--2522"
}

@inproceedings{
LLMSelector,
title={{LLMSELECTOR}: Learning to Select Models in Compound {AI} Systems},
author={Lingjiao Chen and Jared Quincy Davis and Boris Hanin and Peter Bailis and James Zou and Matei Zaharia and Ion Stoica},
booktitle={ICML 2025 Workshop on Collaborative and Federated Agentic Workflows},
year={2025},
url={https://openreview.net/forum?id=NphowWHYJj}
}

@article{candes2009exact,
author = {Cand\`{e}s, Emmanuel J. and Recht, Benjamin},
title = {Exact Matrix Completion via Convex Optimization},
year = {2009},
issue_date = {December  2009},
publisher = {Springer-Verlag},
address = {Berlin, Heidelberg},
volume = {9},
number = {6},
issn = {1615-3375},
journal = {Found. Comput. Math.},
month = dec,
pages = {717–772},
numpages = {56}
}

@article{candes2010power,
author = {Cand\`{e}s, Emmanuel J. and Tao, Terence},
title = {The power of convex relaxation: near-optimal matrix completion},
year = {2010},
issue_date = {May 2010},
publisher = {IEEE Press},
volume = {56},
number = {5},
issn = {0018-9448},
url = {https://doi.org/10.1109/TIT.2010.2044061},
doi = {10.1109/TIT.2010.2044061},
journal = {IEEE Trans. Inf. Theor.},
month = may,
pages = {2053–2080},
numpages = {28}
}

@article{keshavan2010matrix,
author = {Keshavan, Raghunandan H. and Montanari, Andrea and Oh, Sewoong},
title = {Matrix completion from a few entries},
year = {2010},
issue_date = {June 2010},
publisher = {IEEE Press},
volume = {56},
number = {6},
issn = {0018-9448},
url = {https://doi.org/10.1109/TIT.2010.2046205},
doi = {10.1109/TIT.2010.2046205},
journal = {IEEE Trans. Inf. Theor.},
month = jun,
pages = {2980–2998},
numpages = {19}
}

@article{cai2010singular,
author = {Cai, Jian-Feng and Cand\`{e}s, Emmanuel J. and Shen, Zuowei},
title = {A Singular Value Thresholding Algorithm for Matrix Completion},
year = {2010},
issue_date = {January 2010},
publisher = {Society for Industrial and Applied Mathematics},
address = {USA},
volume = {20},
number = {4},
issn = {1052-6234},
journal = {SIAM J. on Optimization},
month = mar,
pages = {1956–1982},
numpages = {27}
}

@misc{li2026llmevaltensor,
      title={LLM Evaluation as Tensor Completion: Low Rank Structure and Semiparametric Efficiency}, 
      author={Jiachun Li and David Simchi-Levi and Will Wei Sun},
      year={2026},
      eprint={2604.05460},
      archivePrefix={arXiv},
      primaryClass={stat.ME},
      url={https://arxiv.org/abs/2604.05460}, 
}

@article{MuSiQue,
  title={{MuSiQue: Multihop Questions via Single-hop Question Composition}},
  author={Trivedi, Harsh and Balasubramanian, Niranjan and Khot, Tushar and Sabharwal, Ashish},
  journal={Transactions of the Association for Computational Linguistics},
  volume={10},
  pages={539--554},
  year={2022},
  publisher={MIT Press One Broadway, 12th Floor, Cambridge, Massachusetts 02142, USA~…}
}

@inproceedings{2WikiMultiHopQA,
    title = {"Constructing A Multi-hop {QA} Dataset for Comprehensive Evaluation of Reasoning Steps"},
    author = "Ho, Xanh  and
      Duong Nguyen, Anh-Khoa  and
      Sugawara, Saku  and
      Aizawa, Akiko",
    editor = "Scott, Donia  and
      Bel, Nuria  and
      Zong, Chengqing",
    booktitle = "Proceedings of the 28th International Conference on Computational Linguistics",
    month = dec,
    year = "2020",
    address = "Barcelona, Spain (Online)",
    publisher = "International Committee on Computational Linguistics",
    url = "https://aclanthology.org/2020.coling-main.580/",
    doi = "10.18653/v1/2020.coling-main.580",
    pages = "6609--6625"
}

@misc{AWS_Quick_Flows,
  author       = {{Amazon Quick Flows}},
  title        = {{Amazon Quick Flows}},
  year         = {2025},
  howpublished = {\url{https://aws.amazon.com/quick/flows/}},
}

@misc{Google_Vertex_Agent_Engine,
  title        = {Vertex AI Agent Engine Overview},
  author       = {{Vertex AI}},
  year         = {2025},
  howpublished = {\url{https://docs.cloud.google.com/agent-builder/agent-engine/overview}},
}

@misc{Azure_AI_foundry,
  author       = {{Azure AI Foundry}},
  author        = {{Azure AI Foundry: Build, Deploy, and Manage AI Applications and Agents}},
  year         = {2025},
  howpublished = {\url{https://azure.microsoft.com/en-us/products/ai-foundry/}},
}

@misc{LangChain,
  title        = {{LangChain}},
  author       = {Chase, Harrison},
  year         = {2022},
  howpublished = {\url{https://www.langchain.com/}},
}

@misc{LLMChemistry,
      title={{LLM Chemistry Estimation for Multi-LLM Recommendation}}, 
      author={Huascar Sanchez and Briland Hitaj},
      year={2025},
      eprint={2510.03930},
      archivePrefix={arXiv},
      primaryClass={cs.LG},
      url={https://arxiv.org/abs/2510.03930}, 
}

@inproceedings{FLORA,
title={{{FLORA}: {GNN}s as Predictors of Agentic Workflow Performances}},
author={Yuanshuo Zhang and Yuchen Hou and Bohan Tang and Shuo Chen and Muhan Zhang and Xiaowen Dong and Siheng Chen},
booktitle={The Fourth Learning on Graphs Conference},
year={2025},
url={https://openreview.net/forum?id=RmlDI5kUxa}
}

@inproceedings{AgenticPredictor,
title={{Multi-View Encoders for Performance Prediction in {LLM}-Based Agentic Workflows}},
author={Patara Trirat and Wonyong Jeong and Sung Ju Hwang},
booktitle={The Fourteenth International Conference on Learning Representations},
year={2026},
url={https://openreview.net/forum?id=7oeKDZsmWp}
}

@misc{ScaffoldedLLMs,
      title={{Scaffolded Language Models with Language Supervision for Mixed-Autonomy: A Survey}}, 
      author={Matthieu Lin and Jenny Sheng and Andrew Zhao and Shenzhi Wang and Yang Yue and Victor Shea Jay Huang and Huan Liu and Jun Liu and Gao Huang and Yong-Jin Liu},
      year={2025},
      eprint={2410.16392},
      archivePrefix={arXiv},
      primaryClass={cs.CL},
      url={https://arxiv.org/abs/2410.16392}, 
}

@misc{Nalar,
      title={Nalar: An agent serving framework}, 
      author={Marco Laju and Donghyun Son and Saurabh Agarwal and Nitin Kedia and Myungjin Lee and Jayanth Srinivasa and Aditya Akella},
      year={2026},
      eprint={2601.05109},
      archivePrefix={arXiv},
      primaryClass={cs.DC},
      url={https://arxiv.org/abs/2601.05109}, 
}

@inproceedings{AFlow,
title={{{AF}low: Automating Agentic Workflow Generation}},
author={Jiayi Zhang and Jinyu Xiang and Zhaoyang Yu and Fengwei Teng and Xiong-Hui Chen and Jiaqi Chen and Mingchen Zhuge and Xin Cheng and Sirui Hong and Jinlin Wang and Bingnan Zheng and Bang Liu and Yuyu Luo and Chenglin Wu},
booktitle={The Thirteenth International Conference on Learning Representations},
year={2025},
url={https://openreview.net/forum?id=z5uVAKwmjf}
}

@inproceedings{EvalGen,
author = {Shankar, Shreya and Zamfirescu-Pereira, J.D. and Hartmann, Bjoern and Parameswaran, Aditya and Arawjo, Ian},
title = {{Who Validates the Validators? Aligning LLM-Assisted Evaluation of LLM Outputs with Human Preferences}},
year = {2024},
isbn = {9798400706288},
publisher = {Association for Computing Machinery},
address = {New York, NY, USA},
url = {https://doi.org/10.1145/3654777.3676450},
doi = {10.1145/3654777.3676450},
booktitle = {Proceedings of the 37th Annual ACM Symposium on User Interface Software and Technology},
articleno = {131},
numpages = {14},
location = {Pittsburgh, PA, USA},
series = {UIST '24}
}

@inproceedings{Prometheus,
title={{Prometheus: Inducing Fine-Grained Evaluation Capability in Language Models}},
author={Seungone Kim and Jamin Shin and Yejin Cho and Joel Jang and Shayne Longpre and Hwaran Lee and Sangdoo Yun and Seongjin Shin and Sungdong Kim and James Thorne and Minjoon Seo},
booktitle={The Twelfth International Conference on Learning Representations},
year={2024},
url={https://openreview.net/forum?id=8euJaTveKw}
}

@article{JMLR_matrixcompletion,
  author  = {Benjamin Recht},
  title   = {A Simpler Approach to Matrix Completion},
  journal = {Journal of Machine Learning Research},
  year    = {2011},
  volume  = {12},
  number  = {104},
  pages   = {3413--3430},
  url     = {http://jmlr.org/papers/v12/recht11a.html}
}

@misc{multiagentllmsystemsfail,
      title={{Why Do Multi-Agent LLM Systems Fail?}}, 
      author={Mert Cemri and Melissa Z. Pan and Shuyi Yang and Lakshya A. Agrawal and Bhavya Chopra and Rishabh Tiwari and Kurt Keutzer and Aditya Parameswaran and Dan Klein and Kannan Ramchandran and Matei Zaharia and Joseph E. Gonzalez and Ion Stoica},
      year={2025},
      eprint={2503.13657},
      archivePrefix={arXiv},
      primaryClass={cs.AI},
      url={https://arxiv.org/abs/2503.13657}, 
}
